\documentclass[sigconf,edbt]{acmart-edbt2021}

\def\BibTeX{{\rm B\kern-.05em{\sc i\kern-.025em b}\kern-.08em
    T\kern-.1667em\lower.7ex\hbox{E}\kern-.125emX}}

\usepackage{booktabs} 
\usepackage{multirow}
\usepackage{tabularx}

\usepackage{pgfplots}

\setcopyright{rightsretained}

\acmDOI{}

\acmConference[arXiv]{arXiv.org}{}{}
\acmYear{2021}

\usepackage{tikz}

\usepackage[outdir=./]{epstopdf}
\usepackage{graphicx}
\usepackage{blindtext}
\usepackage{enumitem}

\usepackage{textcomp}
\usepackage{xcolor}
\usepackage{booktabs} 

\usepackage{amsmath}
\usepackage{algorithm}
\usepackage[noend]{algpseudocode}

\usepackage{varwidth}

\usepackage{wrapfig}
\usepackage{subfigure}
\usepackage{hyperref}
\usepackage{float}

\usetikzlibrary{shapes.arrows}
\usepackage{tikz}

 \usepackage[normalem]{ulem}
 \useunder{\uline}{\ul}{}



\begin{document}

\title{Analysis of key flavors of event-driven predictive maintenance using logs of phenomena described by Weibull distributions
}

\author{Petros Petsinis}
\affiliation{%
  \institution{Aristotle University of Thessaloniki, Greece}
}
\email{petsinispet@gmail.com }

\author{Athanasios Naskos}
\affiliation{%
  \institution{Atlantis Engineering, \\Thessaloniki, Greece}
}
\email{naskos@abe.gr}

\author{Anastasios Gounaris}
\affiliation{%
  \institution{Aristotle University of Thessaloniki, Greece}
}
\email{gounaria@csd.auth.gr}

\renewcommand{\shortauthors}{}

\begin{abstract}
This work explores two approaches to  event-driven predictive maintenance in Industry 4.0 that cast the problem at hand as a classification or a regression one, respectively, using as a starting point two state-of-the-art solutions. For each of the two approaches, we examine different data preprocessing techniques, different prediction algorithms and the impact of ensemble and sampling methods. Through systematic experiments regarding the aspects mentioned above, we aim to 
understand the strengths of the alternatives, and more importantly, shed light on how to navigate through the vast number of such alternatives in an informed manner. Our work constitutes a key step towards understanding the true potential of this type of data-driven predictive maintenance as of to date, and assist practitioners in focusing on the aspects that have the greatest impact.
\end{abstract}

%
%




\maketitle

\section{Introduction}

One of the key characteristics of the 4th Industrial Evolution, broadly known as Industry 4.0, is the wide spreading of Predictive Maintenance (PdM), which is reported to be capable of yielding \emph{``tremendous''} benefits.\footnote{https://www.pwc.nl/nl/assets/documents/pwc-predictive-maintenance-beyond-the-hype-40.pdf}  As such, PdM has evolved to a multi-billion (i.e. \$23.8B) worth market, according to the IoT Analytics industry report \cite{scully2019iot}. PdM has attracted the attention from big companies like IBM\footnote{\url{https://www.ibm.com/services/technology-support/multivendor-it/predictive-maintenance}}, SAP\footnote{\url{https://www.sap.com/products/predictive-maintenance.html}}, SAS\footnote{\url{https://www.sas.com/en_us/software/asset-performance-analytics.html}}, Siemens\footnote{\url{https://new.siemens.com/global/en/products/services/industry/predictive-services.html}}, GE\footnote{\url{https://www.ge.com/digital/iiot-platform}} and ABB\footnote{\url{https://ability.abb.com/customer-segments/marine-ports/remote-diagnostics-predictive-maintenance/}} to smaller companies that focus solely on that field, like Augury\footnote{\url{https://www.augury.com/category/predictive-maintenance/}}, Falkonry\footnote{\url{https://falkonry.com/}}, Sight Machine\footnote{\url{https://sightmachine.com/product/ema/}} and other \cite{scully2019pdm}. 

The goal of PdM is to eliminate machinery downtime and operational costs. Broadly, PdM capitalizes on the benefits stemming from mature techniques in the fields of data analytics, machine learning  and big data management in distributed,  IoT and edge-computing  settings\cite{LU20171}, i.e., it essentially constitutes a specific case of applied data science \cite{Ullman20,Ozsu20}. The main functionality of PdM to achieve its overarching goal is to 
predict the failure of the equipment. To this end, someone can resort to domain knowledge encapsulated in accurate models and/or rules covering the equipment behavior. A second direction, which is leveraged in this work, is driven by the events and log data regarding the equipment operation and maintenance, i.e., it is a data-driven direction. These event logs are processed  to extract patterns of equipment failure automatically. Typically, the event-based PdM techniques rely on methods like data mining, feature selection and machine learning. The main subject of this work is to study, contribute, compare and come to conclusions on two different data-driven PdM approaches that cast the problem of scheduled maintenance into a problem of classification and regression, respectively, while the raw log data comprise a series of event occurrences following a Weibull distribution. This type of distribution is widely used to describe industrial equipment failures  \cite{Lai2006}.

More specifically, we take the state-of-the-art solutions in  \cite{wang2017predictive} and \cite{korvesis2018predictive} as a starting point and explore the impact of employing different model construction, dimensionality reduction, dataset preparation and ensemble techniques. The testing framework is based on the one in \cite{NaskosG19} with appropriate extensions. However, in both solutions the main rationale is the same: firstly, discrete events are continuously monitored and logged, and secondly, before the failure, there are warning events adhering to no crisp patterns (therefore, no sequential pattern mining solutions are effective) and the aim of PdM is to learn to identify such early phenomena. The two pillar solutions of this work, namely \cite{wang2017predictive} and \cite{korvesis2018predictive} come with different approaches regarding the selection of their features and the building of a prediction model (classification vs regression). More importantly, they have been successfully extended to build solutions for PdM in Industry 4.0 \cite{naskos2019event} and detect threat patterns in an IT infrastructure \cite{Bellas20}. However, their application is tricky because it involves decision taking at various design levels, while the performance is sensitive to several parameters. 

In this work, we aim to answer the following questions with a view to deepening the understanding of the afore-mentioned PdM techniques: (1) \emph{``What is the impact on the performance of different prediction techniques compared to the ones originally employed or evaluated?''}, which calls for evaluation of an extensive set of predictors. (2) \emph{``Is the behavior of feature selection techniques correlated with the exact prediction technique employed''? }, which aims to understand whether the feature selection and prediction techniques can be addressed independently in a  divide-and-conquer rationale to reduce the search space complexity. (3) \emph{``Which are the most important dataset attributes that affect the technique choices?''}, which calls for experimenting with various dataset types.
(4) \emph{``Does the manipulation of the training set to deal with class imbalance and small training  sizes can yield benefits?}; to answer this question we explore several alternatives. (5) \emph{``What is the impact of ensemble solutions?''}, where we build on top of the single-predictor results on the quest for building an efficient combination, something that we show that it is challenging. The main contribution of this work is that it provides insightful answers to these questions, and thus constitutes a key step towards understanding the true potential of event-driven predictive maintenance as of to date, when log entries of different types follow a Weibull distribution, and assist practitioners in focusing on the aspects that have the greatest impact. All the code is publicly available.\footnote{ \url{https://github.com/petsinis/JupyterNotebookPdM}}

We provide the background about the main methodologies around which our analysis revolves, along with related work in the subsequent section.  The datasets and the rationale of experimental result assessment are described in Section \ref{sec:exp}. The next section comprises the first main part of our contribution: the alternative feature selection and model building techniques along with their experimental evaluation and key observations. In Section \ref{sec:train}, we explore the impact of training set generation and, in Section  \ref{sec:ensemble}, we deal with ensemble solutions. We conclude in Section \ref{sec:concl} with our final remarks.

\section{Related Work and Background}
\label{sec:back}

In this section, we first present the related work and then we give the background details regarding the two main methodologies considered in our experimental analysis. 

\subsection{Related Work}

Data-driven techniques, where the data refer to past events, commonly in the form of log entries, are widely used in PdM.  \cite{korvesis2018predictive} is a key representative of the state-of-the-art, where the unnderlying problem solved is a regression one. Another event-based approach is presented in \cite{sipos2014log}, where historical and service data from a ticketing system are combined with domain knowledge to train a binary classifier for the prediction of a failure. As in several similar works, a feature selection \cite{bach2008bolasso} and an event amplification technique is used to enhance the effectiveness of the SVM-based classifier. Event-based analysis, based on event and failure logs, is also performed in \cite{wang2017predictive}, where it is assumed that the system is capable of generating exceptions and error log entries that are inherently relevant to significant failures. The proposal in \cite{wang2017predictive} relies on pattern extraction and similarity between patterns preceding a failure, while emphasis is posed on feature selection. 

Data-driven PdM is also related to online frequent episodes mining;
research works \cite{ao2015online} and \cite{li2018once} propose techniques in this topic. The key strength of \cite{ao2015online} is that it can apply to an online (i.e. streaming) scenario. \cite{li2018once} further improves upon it through providing solutions for the case where the event types is unbounded.
Complex-event processing (CEP) \cite{KS18a} is also a technology that enables PdM enforcement after warning sequential patterns have been extracted. 

An  overview of the the data-driven PdM is presented in works such as \cite{CarvalhoSVFBA19,8420112}, where 
the authors list techniques for data preprocessing, feature selection, data storing, model building and validation, fault detection and prediction and estimation of remaining useful life. However, each application comes with its unique characteristics that affect the corresponding design \cite{korvesis2018predictive}.
In the remainder of this section, we further elaborate on the solutions in \cite{wang2017predictive} and \cite{korvesis2018predictive}, which are deemed as key representatives of classification- and regression-based event-driven PdM, respectively. An early comparison of these techniques has appeared in \cite{NaskosG19}; the main difference is that in this work, we investigate several alternatives using each of these techniques as a basis (i.e., we keep the main rationale but we explore different techniques in the way the prediction model is built) and we thoroughly assess their efficiency, while previously, the initial proposals were compared without exploring alternatives.

\subsection{A classification-based methodology}

The first methodology considered is characterized by two main features: firstly, it casts the problem as a binary classification one, and secondly, it follows a specific approach to expanding the feature space.  Although it has been proposed in  \cite{wang2017predictive} for   ATM failures,  it is generic 
and thus can be applied to other predictive maintenance scenarios, provided that the data comprise error logs similar to those in the original proposal.

 \begin{table}[tb!]
 \begin{center}
 \begin{tabular}{|c|p{0.6\columnwidth}|}
   \hline
 Parameter &  Description  \\ \hline
  \it{X}  & number of sub-windows inside an $OW$\\ \hline
  \it{M} & length (number of days) of a sub-window \\ \hline
  \it{Y} & length (number of days) of a $PW$  \\ \hline
  \it{Z} & length (number of days) of a $BW$  \\ \hline
 \end{tabular}
 \caption{Window-related parameters.}
 \label{tab:atm-params}
 \end{center}
 \end{table}

The event logs are mapped to time segments; in our scenario, the finest level of granularity of such segments is a day as discussed in more detail in Section \ref{sec:exp}.  The time segments are then grouped into (sliding) windows.  For each prediction, the segments in a
fixed-length observation window ($OW$) are considered and the prediction refers to a subsequent window, the  prediction window ($PW$).
An $OW$ is in turn divided into $X$ sub-windows of length $M$. From each of them, several basic features are exported, which will eventually constitute several of the new features of the $OW$.

During training, for each $OW$, a feature vector is produced, the label of which is given through its corresponding $PW$. More specifically, if $PW$ contains a 
day on which the equipment failed, the $OW$ vector receives a true label; otherwise the label is 0.  To account for real-world scenarios, where failures very close to a correct prediction cannot be practically prevented, there is a third type of 
 fixed interval of days, referred to as the buffer window ($BW$) in between $OW$ and $PW$.  For example, assume that $OW$, $PW$ and $BW$ are 3, 2 and 1 days, respectively. This corresponds to a scenario where predictions are made taking into account the log entries of the 3 most recent days and refer to forthcoming failures in a period from more than 1 day and less than 3 days ahead.
 Table \ref{tab:atm-params} summarizes the window-related parameters. 

Five types of features comprise the final vector for each $OW$ as follows.
\begin{enumerate}
 \item \emph{Basic features:} a frequency vector for each type of error/log in a sub-window within $OW$. Overall $X$ such vectors are produced, which are concatenated. 
 \item \emph{Advanced statistical features:} a statistics vector that contains, for each logged error type the minimum, maximum, and mean  
distance of the error type within the  $OW$ from the beginning of the  $PW$; also the mean distance (and its standard deviation) 
between occurrences  of the same fault types in the $OW$ are considered.
\item \emph{Failure-similarity features:} The Jaccard similarity between this $OW$ and a reference $OW$ with a positive label.
\item \emph{Pattern-based features:} this is a binary vector of patterns of different types of error based on applying association rule mining.
\item \emph{Equipment features:} this includes metadata regarding the specific equipment, such as model, date of installation, and so on. 
\end{enumerate}

In our comparison, we employ the first three types of features. In the original proposal in \cite{wang2017predictive}, several feature selection techniques, such as ReliefF, and classifiers, such as XGBoost, are investigated, In this work, we consider both the best performing techniques in \cite{wang2017predictive} and additional solutions, such as nearest neighbors and neural network-based ones, as detailed in Section \ref{sec:class}, and we discuss the cases in which each alternative behaves better.

\subsection{A regression-based methodology}

Instead of casting the problem as a (binary) classification one, another approach is to cast it as a regression one aiming to reduce the prediction error where each prediction gets a real value, as proposed in \cite{korvesis2018predictive}  to solve the event-based PdM problem in the aviation industry.

In this approach, the fault event logs are grouped in episodes. An episode begins with the first event that occurs after the occurrence of the main failure event that we aim to predict (called target event) and ends with the last event before the occurrence of the next target event.
Contrary to the classification-based methodology that focuses on feature generation, this methodology emphasizes more on log pre-processing. More specifically, the following pre-processing steps are considerd:
\begin{enumerate}
  \item \emph{Deletion of rare events:} if the occurrence of a fault type event is rare compared to that of the target event, then it is considered unrelated to it, so it is not useful and can be removed.
  \item \emph{Deletion of frequent events:} if the occurrence of a fault type event is very frequent, then it is also considered not related to the target event, so it is not useful and can be removed.
  \item \emph{Dealing with multiple appearances:} in some cases, we do not care how many times an event occurred during a time segment. In this methodology, the occurrence vector corresponding to each time period and fed to the classifier is a binary one indicating only  whether an event has appeared at least once in this period.

  \item \emph{Dealing with continuous appearances:} in some cases the appearance of a fault type event in  consecutive time segments 
  can be due to the fact that no action was taken because, for example, of a low priority. To avoid inserting unnecessarily event occurrences,  in the case of  appearances of the same fault type in the log in multiple consecutive time segments, only the first appearance is kept.   
\end{enumerate}
 
\begin{figure}
\begin{tikzpicture}[declare function={sigma1(\x)=1/(1+exp(0.9*(14-\x-7)));
sigma2(\x)=1/(1+exp(0.9*(14-\x-10)));
sigma3(\x)=1/(1+exp(0.4*(14-\x-7)));}]
\begin{axis}
[
    xmin=0,
    xmax=14,
    axis x line=bottom,
    ytick={0,.5,1},
    ymax=1,
    axis y line=middle,
    samples=100,
    domain=0:14,
    legend style={at={(1,0.275)}}     
]
    \addplot[blue,mark=*]   (x,{sigma1(x)});
    \addplot[red,mark=x]   (x,{sigma2(x)});
    \addplot[green,mark=square]   (x,{sigma3(x)});
    \legend{${\sigma1(x): m=7, s=0.9}$,${\sigma2(x): m=10, s=0.9}$,${\sigma3(x): m=7, s=0.4}$}
\end{axis}
\end{tikzpicture}
\caption{Impact of midpoint and steepness on the sigmoid function (the target event is on time point 14).}
\label{fig:sigmoid}
\end{figure}
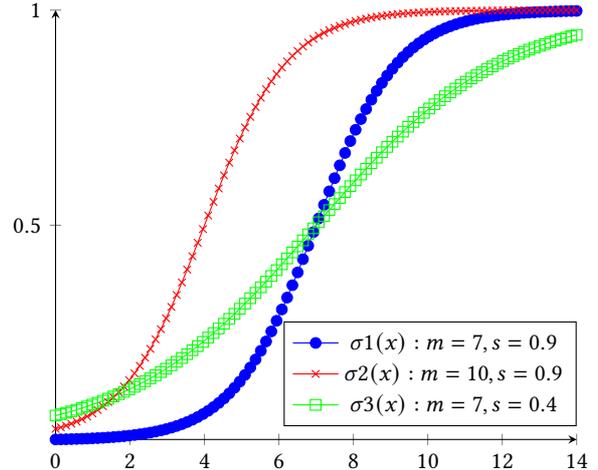

 Overall, each time segment in an episode is described by a binary vector with size equal to the number of different event types in the log. The value associated with this vector during training is derived from a risk function, which encapsulates the rationale that the closer to the target event a set of event types appear, the more they constitute warning signs of a subsequent failure of key equipment. This is naturally quantified with the help of a sigmoid function, as shown in Figure \ref{fig:sigmoid}. The values of this function are bounded in the (0,1) range.
 There are two main parameters, the midpoint $m$ and the steepness $s$. The former defines the distance in number of days from the target event that the sigmoid takes the value of 0.5. The $s$ parameter defines how steep the curve is; the higher its value, the less significant the sets of events before the midpoint become, and similarly, the more significant after the midpoint. 

In the original implementation in \cite{korvesis2018predictive}, the superiority of a Random Forest regression solution upon SVM was shown; here we explore further alternatives, keeping the pre-processing steps as defined above.

%

\section{Experimental setting}
\label{sec:exp}
 
There are several datasets used in PdM as reported in \cite{CarvalhoSVFBA19}. Also, NASA maintains a repository for prognostic datasets.\footnote{\url{https://ti.arc.nasa.gov/tech/dash/groups/pcoe/prognostic-data-repository/}} However, these datasets contain sensor data time series rather than partially ordered sequences of discrete events, which is our focus. To overcome this problem, we re-engineer in our code implementation the event generator in \cite{NaskosG19}, which is publicly available. We first describe the datasets used and then the assessment methodology.
 
\subsection{Datasets}

The data used in this work is a set of timestamped event logs. They are synthetically generated in a manner that resembles real-world applications, such as the one in \cite{korvesis2018predictive}. An example is shown in Table \ref{tab:log}, where the time granularity is complete days. On each day, several events may occur, which yields a partially ordered sequence. Note that the time granularity of complete days does not refer to the frequency of monitoring the status of industrial equipment continuously in a modern Industry 4.0 setting, where, typically, several measurements are taken every second, but to the frequency of recording higher-level   incidents, a (very) small portion of which \emph{may} be relevant to a target event of high importance. Nevertheless, the described approach can be trivially modified to apply to any time granularity. 
 
\begin{table}[tb!]
\begin{center}
 \begin{tabular}{|c c |} 
 \hline
 Timestamps & EventId \\  [0.5ex] 
 \hline \hline
  2020-01-01 & 1  \\ 
 \hline
 2020-01-01 & 22  \\
 \hline
 2020-01-02 & 12  \\
 \hline
 2020-01-02 & 7  \\
 \hline
 2020-01-02 & 12  \\ 
 \hline
2020-01-03 & 7  \\ 
 \hline
2020-01-04 & 3  \\
 \hline
... & ...  \\ 
\end{tabular}
\caption{An example event log}
\label{tab:log}
\end{center}
\end{table}

The main parameters of the generator in \cite{NaskosG19} are provided in Table \ref{tab:dg-params}. In each synthetic dataset, there are \emph{ft} fault/event types. Each one of them follows a Weibull distribution, since this type of distribution can describe several types of equipment failures \cite{Lai2006}.
One event plays the role of the target event. The rationale of the data generator is to artificially inject event patterns before the occurrence of the target event, and then test the capability of the classifiers to detect such patterns and become capable of predicting the target event.

 \begin{table}[tb!]
 \begin{center}
 \begin{tabular}{|c|p{0.6\columnwidth}|}
   \hline
 Parameter &  Description  \\ \hline
  \it{ft}  & number of different event/fault types \\ \hline
  $s_{tr}$ & size of training dataset (in days)  \\ \hline
  $s_{te}$ & size of testing dataset (in days) \\ \hline
  \it{pl} & pattern length  \\ \hline
  $min_t/max_t$ & min./max. distance (in days) of the last pattern event from the target event \\ \hline
  $min_p/max_p$ & min./max. distance (in days) between pattern events  \\ \hline
    $min_f/max_f$ & min./max. pattern event forms \\ \hline
  \it{pc} & pattern clarity  \\ \hline
  \it{pps} & the percentage of the missing events in the distorted patterns \\ \hline
 \end{tabular}
 \caption{Main dataset generator parameters from \cite{NaskosG19}}
 \label{tab:dg-params}
 \end{center}
 \end{table}

A key feature of the data generator is that the injected patterns preceding a target event can be perturbed using several tuning parameters to imitate the fact that, typically, no crisp patterns, which can be more easily detected, appear. More specifically, there may be no full pattern before a target event instance; moreover, a pattern may not always be followed by a target event. The extent to which these two behaviors appear in the logs are configurable through 
the \emph{pc} and \emph{pps} parameters. The clarity percentage (captured by \emph{pc})  defines the percentage of partial patterns, i.e. pattern instances that are missing some of the events of the original pattern. The partial pattern size (\emph{pps}) defines the percentage of the missing events.  For example, setting $pc=0.9$ and $pps=0.5$ enforces 10\% of the pattern instances to include only the half of the events of the original pattern, which is of size \emph{pl}.  In addition, in 10\% of the injected patterns, no target event follows.

Moreover, in a pattern comprising \emph{pl} elements, each of the elements may correspond to multiple event types, captured by the parameters $min_f$ and $max_f$.  This corresponds to the situation, where there are families of faults (event types) that might give early warning for an upcoming  target event. Finally, the pattern elements may be shuffled. Overall, the combination of the \emph{pc}, \emph{pps}, \emph{shuffle}, $min_f$ and $max_f$ parameters allows us to create synthetic datasets, where it is challenging to detect and create models regarding warning event sets.


\begin{table*}[tb!]
\begin{center}
\begin{tabular}{|c|c|c|c|c|c|c|c|c|c|c|c|c|c|}
  \hline
Dataset & $ft$ & $shuffle$   & $pl$ & $min_f$ & $max_f$ & $s_{tr}$ & $s_{te}$ & $min_t$ & $max_t$ & $min_p$ & $max_p$ & $pc$ & $pps$ \\ \hline
DS1 & \multirow{ 2}{*}{150} & \multirow{ 2}{*}{no} &  6 & 1 & 3 & \multirow{ 2}{*}{1094} & \multirow{ 2}{*}{730} & \multirow{ 2}{*}{1} & \multirow{ 2}{*}{5} & \multirow{ 2}{*}{1} & \multirow{ 2}{*}{2} & \multirow{ 2}{*}{90\%} & \multirow{ 2}{*}{50\%} \\  \cline{1-1} \cline{4-6}
DS2 & & &  4 & 3 & 4 & & & & & & & & \\ \hline
DS3  & \multirow{ 2}{*}{1500} & \multicolumn{12}{|c|}{same as DS1} \\ \cline{1-1} \cline{3-14}
DS4  & & \multicolumn{12}{|c|}{} \\ \cline{1-3}
DS5  & \multirow{ 2}{*}{150} & \multirow{ 2}{*}{yes} & \multicolumn{11}{|c|}{same as DS2} \\ \cline{1-1} 
DS6  & & & \multicolumn{11}{|c|}{} \\ \hline
\end{tabular}
\caption{Dataset generator parameters taken from \cite{NaskosG19}; $DS1$ to $DS5$ contain approx. 50 target events, whereas $DS6$ contains 25 target events.}
\label{tab:params}
\end{center}
\end{table*}

Overall, in the evaluation, 6 types of synthetic datasets are generated, as shown in Table \ref{tab:params}. The types of the datasets are the same as in \cite{NaskosG19} but the random instances of each dataset type are different.  The datasets correspond to a period of 5 years and are split into 3 years of training logs and 2 years of test logs. 
The distance of the warning pattern from the target event ranges from 1 to 5 days, while the distance between the events of the pattern ranges from 1 to 2 days. The pattern clarity is set to 90\%, while the partial patterns include only the half of the events of the full pattern (i.e. $pps=50\%$).  These datasets comprise approximately 50 target events (on average, one target event in 36 days) apart from the last dataset type, which comprises only 25 target events.
There are two main dataset types, namely, DS1 and DS2 and the other datasets (i.e., DS3-6) are slightly altered versions of the main ones. More specifically, DS3 and DS4 have 10X more event types than DS1 and DS2, respectively. DS5 and DS6 include shuffled patterns compared to DS2, while, as already mentioned above, DS6 includes fewer target events. In the experiments, ten instances of each dataset are produced.

\subsection{Evaluation Methods}

The test set is divided into episodes, where each episode ends with a target event. There are three main time intervals within an episode that impact on the evaluation: 
\begin{enumerate}
\item The \emph{Correct Prediction} period, in which the prediction of an equipment's failure is made in a timely manner. The predictions falling into this period as counted as True Positives (TP). 

\item The \emph{Early Prediction Period}, which starts just after the occurrence of the previous failure (target event) and ends just before the correct prediction period. The predictions for failure during this period are considered to be premature and unnecessary. They are counted as False Positives (FP)

\item The \emph{Repair period}, which represents the days needed to repair the fault before it manifests itself. If equipment's failure is predicted  within the repair period, then it is ignored. 
\end{enumerate}

%

\begin{table}[tb!]
\resizebox{\columnwidth}{!}{%
\begin{tabular}{|l|l|l|} \hline
Case& \multicolumn{2}{l|}{Description} \\ 
\hline\hline
True Positive (TP)& \multicolumn{2}{p{5.5cm}|}{\raggedright Set to 1 if at least one failure prediction made during the Correct Prediction Period; otherwise, it  is set to 0.}\\
\hline
False Negative (FN) & \multicolumn{2}{p{5.5cm}|}{\raggedright   Set to1 if no failure prediction made during the Correct Prediction Period; otherwise, it  is set to 0.}\\
\hline
False Positive (FP) & \multicolumn{2}{p{5.5cm}|}{\raggedright  The amount of failure predictions during  the Early Prediction Period.}\\
\hline
True Negative (TN)& \multicolumn{2}{p{5.5cm}|}{\raggedright The amount of non-failure predictions during the Early Prediction Period.}\\
\hline
Ignored& \multicolumn{2}{p{5.5cm}|}{\raggedright All failure predictions during the Repair Period.}\\
\hline
\end{tabular}}
\caption{Description of prediction cases within an episode.}
\label{tab:cases}
\end{table}

Table \ref{tab:cases} presents all the prediction cases within an episode. Note that there can be up to 1 TP or FN per episode, but arbitrary numbers of FPs and TNs.  In all the experiments, we have used the F1 metric to evaluate the results:
\[ F1 \; Score=2\frac{{Precision*Recall}}{{Precision+Recall}} \] where
\(Precision=\frac{{TP}}{{TP+FP}}\) and \(Recall=\frac{{TP}}{{TP+FN}} .\)

The reason behind employing F1 is that it balances recall and precision. Recall is important, since it reflects the number of failures predicted. However, it is trivial to create a dummy predictor that always votes for failure. Therefore, we need to balance recall with precision, exactly as F1 targets. 

\begin{figure}[tb!]
\centering
\begin{tikzpicture}
\draw (0,0) --(0,0.4) --(2,0.4)  --(2,0) --(0,0)--(2,0)  node[pos=0.5,above,scale=.65]{Early Prediction Period}--(2,0.4)--(5,0.4)--(5,0) --(2,0)  node[pos=0.5,above,scale=.75]{Correct Prediction Period}--(5,0)--(5,0.4) --(7,0.4)--(7,0) --(5,0)  node[pos=0.5,above,scale=.75]{Repair Period};
\draw[blue, dashed, very thick]
(7,0)--(7,-0.825)--(3,-0.825)--(3,0);
\draw[blue, dashed, very thick]
(5.5,0)--(5.5,-0.55)--(1.5,-0.55)--(1.5,0);
\draw[blue, dashed, very thick]
(4,0)--(4,-0.275)--(0,-0.275)--(0,0);
\filldraw 
(-0.01,0) circle (1pt) node[align=left,   below] {Failure};
\filldraw 
(7.01,0) circle (1pt) node[align=right,   below] {Failure}--
(1,0) circle (1pt) --
(2,0) circle (1pt) --
(3,0) circle (1pt)--
(4,0) circle (1pt)--
(5,0) circle (1pt) --
(6,0) circle (1pt)--
(7,0) circle (1pt) --
(0.5,0) circle (1pt) --
(1.5,0) circle (1pt) --
(2.5,0) circle (1pt) --
(3.5,0) circle (1pt)--
(4.5,0) circle (1pt)--
(5.5,0) circle (1pt) --
(6.5,0) circle (1pt);


\end{tikzpicture}
\caption{Example of dividing a sequence of fourteen days to 3 windows, using moving step of 3 days.} \label{fig:M1}
\end{figure}
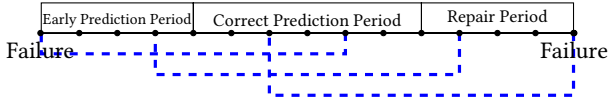

In the classification-based technique, the sequence of days is divided into windows ($OW$s) using a moving step of predefined length (see also Figure \ref{fig:M1}). One prediction is made for each different window slide. The interval in which the last day of the window falls determines the type of forecast (TP, TN, FP, FN), as explained above.

In the regression-based technique as initially described in \cite{korvesis2018predictive}, the predictions were made on a daily basis. Since this technique assigns, in each prediction, a risk value, a threshold is specified. If the risk value is below a threshold, then this is deemed as a non-failure prediction. If it is equal to or greater than the threshold, a failure alarm is raised.  In order for the comparison of the two techniques to be fair, the predictions of the regression-based technique are made for each group of $N$ days, where $N$ is equal to the size of the $OW$. Also, the training is done using a moving window. As previously,  the risk value received on the last day of the $N$ days will be used during training; also the period in which this day falls determines the manner the prediction outcome is assessed.

\section{Impact of classifiers and dimensionality reduction}
\label{sec:class}
 
In this section, we first present the alternatives for model building for the two main approaches under investigation. These alternatives extend the ones included originally in the proposals in \cite{wang2017predictive} and \cite{korvesis2018predictive}. Then we present the alternatives for dimensionality reduction that are examined. For each of the alternatives introduced, we present their main settings.\footnote{These setting have been found after experimentation that is not presented in this work due to space limitations.} Finally, we conduct a thorough experimental analysis.

\subsection{Classification and Regression Predictors}

\subsubsection{K-Nearest Neighbors (used for both techniques)}
One of the earliest classifiers, but still widely used today, is the \emph{K-Nearest Neighbors} (KNN). It belongs to the category of lazy classifiers, since it does not rely on model building. There exist several flavors; in our work, KNN  classifies an unknown sample into the class of the largest majority among the $k$ closest samples. As such, it comprises two main configuration parameters. The first one refers to the definition of the distance. In our work, the distance metric employed is the Euclidean one. The second tuning knob is the value of $k$, which, here, is set in the range of [1,5]. 

\subsubsection{Neural Networks (used for both techniques)}
Another broad category of classifiers is \emph{(Artificial) Neural Networks} (NNs), which are inspired by the manner the brain operates based on biological neural networks. Briefly,  the purpose of a NN is to minimize a cost or loss function, which depends on the problem in hand. For example, for the classification problem, a NN is typically trained so that, by entering and processing the training examples, it minimizes the wrong predictions (maximizes the correct predictions). In this regard, the classification problem is a binary one, and we employ the 
binary crossentropy loss function. 
In the regression-based technique, minimizing the mean square error had the best reuslt and this is used for the experiments presented later. The error can be any metric distance between the predicted and the actual value, such as the absolute difference between the risk values. 
The optimization is based on stochastic gradient descent, and more specifically, on Adam \cite{KingmaB14}, with  fixed learning rate=0.001.

As holds for all classifiers, a key objective for NNs is generalization. To this end, many NN architectures have been proposed, three of them were examined in the present work.

\paragraph{Multilayer Perceptron (used for both techniques)} Multilayer Perceptron  (MLP) is one of the most well-known and widely used NN architectures consisting of at least one hidden level and characterized by a high degree of interconnection between the neurons of two successive levels. The NN are further configured as follows. Each neuron has a rectifying activation function (ReLU) \cite{ramachandran2017searching}.  For both techniques, we have used  3-layer MLPs with 128, 64 and 128 units, respectively and fixed dropout of 25\% in the last two layers.  For both the binary classification and the reggression problem, there is only one output neuron.

\paragraph{Long Short Term Memory (used for both techniques)} Long Short Term Memory (LSTM) is a Recurrent Neural Network (RNN) architecture  that has feedback connections. As such,  
a RNN is just a multi-layer NN where there are connections between higher-level outputs and inputs of the same or lower levels. As a result, the entry of a neuron at time $t$ does not depend only on the input data at $t$, but also on the previous values $t-1$, $t-2$, and so on, so that the system acquires memory. 
LSTMs can be effectively applied to time series classification problems like the ones in our setting. However, the downside is that they need a relatively larger amount of training data. 
For this study, the LSTM architecture has only one hidden layer with 128 neurons.

\begin{figure}[tb!]
  \centering
  \includegraphics[width=0.75\columnwidth]{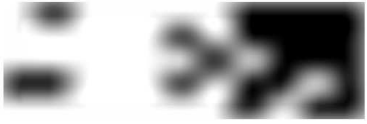}
  \caption{Treating an observation window as an image.}
  \label{fig:cnn}
\end{figure}

\paragraph{Convolutional Neural Network (used for both techniques)} Convolutional Neural Network (CNN) is an extension to MLPs. It is widely used in image detection as it can effectively detect patterns in images using filters. The fixed scheme of CNNs that is used comprises two convolution layers followed by an average pooling layer, with dropout 20\% just before the output level;  the kernel size is set to 3 and ReLU is employed  for the activation function. As in the case of LSTMs, the input examples are not a vector but a set of vectors. We
can imagine training samples not as simple file logs but as images. If one considers that the images are two-dimensional arrays with numbers depending on the color representation (e.g., 0 for black and 1 for white), we can think our samples as images with a view to benefiting from a NN technique excelling in image analysis, such as CNN (see also Figure \ref{fig:cnn}).

\subsubsection{XGBoost (used for both techniques)} It is an implementation of gradient boosting \cite{chen2016xgboost}, which minimizes the average square error using gradient descent.
It starts with the creation of an initial tree model that predicts the average of the labels of each training sample.
It continues with the creation of a new tree model to predict the errors of the previous tree model. This process continues until a certain number of models (i.e., an ensemble) are created or until it converges.

\subsubsection{Random Forest (used for both techniques)} This also belongs to ensemle methods but follows a bagging approach. Random Forest (RF) is a meta estimator that fits a number of decision tree classifiers on various samples of the dataset and uses averaging to improve the predictive accuracy and mitigate over-fitting. It is earlier than XGBoost.

\subsubsection{Partial Least Squares (used for regression)} Partial Least Squares (PLS) is a method of predicting dependent values from a set of observations. In our tests, we have used a PLS version that encapsulates 	 singular value decomposition (SVD) to 
decompose the matrices  of the initial data  into components. 

\subsection{Dimensionality Reduction Algorithms}

Dimensionality reduction is particularly relevant in our setting given that very few event types have actually predictive power; the vast majority  of event types is irrelevant to the failure target events. By pruning and decreasing the search space, the classifier performance can be significantly improved. In this work, we have investigated two techniques to this end.

\subsubsection{ReliefF \cite{kononenko1997overcoming}}
This algorithm implements an iterative process that takes as input training examples along with their corresponding labels and exports a vector with quality estimates for each attribute. The evaluation of each feature is done using two concepts based on the $k$ nearest hits and misses, respectively.

\subsubsection{PCA} The Principal Component Analysis allows to  transform a high-dimensional dataset to a low-dimensional one while losing as little information as possible. In our application of PCA, the new space is enforced after subtracting the mean in each dimension, i.e., we center the dataset in the beginning of the axes.

\subsection{Evaluation of Classification Predictors}

In this subsection, we explore different classifiers for the class-ification-based technique inspired by \cite{wang2017predictive}.
For each of the 6 dataset types in Section \ref{sec:exp}, we have created 10 random instances, which remain the same through all experiments to allow for an apples-to-apples comparison. We have also chosen 4 main settings for the classification-based technique, as described below:

\begin{enumerate}
\item \textbf{A}, where
the period $Y$ within which the prediction can be made is 16 days (approximately 2 weeks) and we set $X=M=4$, so that $X \cdot M=Y$, i.e., the observation window is of the same size as the prediction window. Also, there is no buffer window, i.e., $Z=0$.
\item \textbf{B}, where we shrink $Y$ to 6 days, and we set $X=3$ and $M=2$.
\item \textbf{C}, which is similar to \textbf{A} but with $Z=3$.
\item \textbf{D}, which is similar to \textbf{B} but with $Z=3$.
\end{enumerate}
The length of the step in each case is defined as half the length of the prediction window in order for each day to be considered in 2 predictions.

For each algorithm, several configurations are tested and the best performing tuning on average is chosen. For PCA and ReliefF to find the best performing selected features that maximize the score, we repeatedly reduce the dimensions by a percentage of 20\%.
Table \ref{tab:classDS1} presents
the mean F1 score of all combinations of classifiers and dimensionality reduction techniques. At the bottom of the table, we show two naive classifiers, one random and one always raising an alarm. For each classifier, the best performing dimensionality reduction setting is underlined. The overall best approach on average is in bold.

\begin{table}[tb!]
\begin{tabular}{|c|c|c|c|c|c|}
\hline
\multicolumn{2}{|c|}{Setting}                                                                    & A               & B                  & C                   & D                   \\ \hline
Algorithm                                           & Feat.   Sel.                           & \multicolumn{4}{c|}{Mean F1 score in 10   DS1 instances}                                          \\ \hline
\multirow{3}{*}{XGBoost}                            & None                                          & 0.685                & 0.225                & 0.600                & 0.321                \\ \cline{2-6} 
                                                    & PCA                                           & 0.654                & 0.315                & 0.670                &  0.387         \\ \cline{2-6} 
                                                    & ReliefF                                       & {\ul\textbf{0.752}}          &{ \ul\textbf{0.613}} &   {\ul\textbf{0.756}}          & {\ul\textbf{0.609}} \\ \hline               
\multirow{3}{*}{RF}                                 & None                                          & 0.181                & 0.000                & 0.034                & 0.000                \\ \cline{2-6} 
                                                    & PCA                                           & 0.504                & 0.059                & 0.501                & {\ul 0.259}          \\ \cline{2-6} 
                                                    & ReliefF                                       & {\ul 0.697}          & {\ul 0.244}          & {\ul 0.587}          & 0.203                \\ \hline
\multirow{3}{*}{KNN}                                & None                                          & 0.543                & 0.066                & 0.499                & 0.046                \\ \cline{2-6} 
                                                    & PCA                                           & 0.671                & 0.269                & {\ul 0.672}          & {\ul 0.420} \\ \cline{2-6} 
                                                    & ReliefF                                       & {\ul 0.698}          & {\ul 0.273}          & 0.660                & 0.326                \\ \hline
\multirow{3}{*}{MLP}                                & None                                          & 0.606                & 0.150                & 0.576                & 0.259                \\ \cline{2-6} 
                                                    & PCA                                           & {\ul 0.656}          & 0.280                & {\ul 0.670}          & {\ul 0.414}          \\ \cline{2-6} 
                                                    & ReliefF                                       & 0.620                & {\ul 0.282}          & 0.667                & 0.372                \\ \hline
\multirow{3}{*}{LSTM}                               & None                                          & 0.532                & 0.204                & 0.475                & 0.236                \\ \cline{2-6} 
                                                    & PCA                                           & 0.623                & 0.322                & 0.574                & {\ul 0.370}          \\ \cline{2-6} 
                                                    & ReliefF                                       & {\ul 0.644}          & {\ul 0.330}          & {\ul 0.608}          & 0.362                \\ \hline
\multirow{3}{*}{CNN}                                & None                                          & 0.593                & 0.204                & 0.589                & 0.291                \\ \cline{2-6} 
                                                    & PCA                                           & {\ul 0.612}          & {\ul 0.346}          & 0.598                & 0.366                \\ \cline{2-6} 
                                                    & ReliefF                                       & 0.610                & 0.331                & {\ul 0.642}          & {\ul 0.376}          \\ \hline
\multicolumn{2}{|c|}{Random Classifier}                                                             & 0.520                & 0.223                & 0.543                & 0.275                \\ \hline
\multicolumn{2}{|c|}{All True Classifier}                                                           & 0.464                & 0.176                & 0.500                & 0.190                \\ \hline
\end{tabular}
\caption{Mean F1 score of classification predictors for DS1.}
\label{tab:classDS1}
\end{table}

\begin{table}[tb!]
\begin{tabular}{|c|c|c|c|c|c|}
\hline
\multicolumn{2}{|c|}{Setting}                                                                    & A               & B                  & C                   & D                   \\ \hline
Algorithm                                           & Feat.   Sel.                           & \multicolumn{4}{c|}{Min F1 score in 10   DS1 instances}                                          \\ \hline
\multirow{3}{*}{XGBoost}                            & None                                          & 0.605                & 0.12                & 0.474                & 0.19                \\ \cline{2-6} 
                                                    & PCA                                           & 0.615                & 0.233                & 0.571                &  0.338         \\ \cline{2-6} 
                                                    & ReliefF                                       & {\ul\textbf{0.653}}          &{ \ul\textbf{0.5}} &   {\ul\textbf{0.676}}          & {\ul\textbf{0.588}} \\ \hline               
\multirow{3}{*}{RF}                                 & None                                          & 0.000                & 0.000                & 0.000                & 0.000                \\ \cline{2-6} 
                                                    & PCA                                           & 0.435                & 0.000                & 0.316                & {\ul 0.129}          \\ \cline{2-6} 
                                                    & ReliefF                                       & {\ul 0.562}          & {\ul 0.129}          & {\ul 0.435}          & 0.129              \\ \hline
\multirow{3}{*}{KNN}                                & None                                          & 0.4                & 0.000                & 0.375                & 0.000                \\ \cline{2-6} 
                                                    & PCA                                           &{\ul 0.586}                & {\ul 0.211}                & {\ul 0.588}          & {\ul 0.355} \\ \cline{2-6} 
                                                    & ReliefF                                       &  0.583          & 0.2          & 0.565                & 0.304                \\ \hline
\multirow{3}{*}{MLP}                                & None                                          & 0.409                & 0.045                & 0.449                & 0.000                \\ \cline{2-6} 
                                                    & PCA                                           & {\ul 0.571}          & {\ul 0.214 }               & 0.548          & {\ul 0.311}          \\ \cline{2-6} 
                                                    & ReliefF                                       & 0.543                & 0.213          & {\ul 0.6}               & 0.259                \\ \hline
\multirow{3}{*}{LSTM}                               & None                                          & 0.429                & 0.111                & 0.381                & 0.111                \\ \cline{2-6} 
                                                    & PCA                                           & 0.444                & 0.214                & 0.44                & 0.279          \\ \cline{2-6} 
                                                    & ReliefF                                       & {\ul 0.571}          & {\ul 0.291}          & {\ul 0.543}          & {\ul 0.314}                \\ \hline
\multirow{3}{*}{CNN}                                & None                                          & 0.451                & 0.08                & 0.44                & 0.17                \\ \cline{2-6} 
                                                    & PCA                                           & {\ul 0.545}          & 0.224          & 0.508                & 0.299              \\ \cline{2-6} 
                                                    & ReliefF                                       & 0.54                & {\ul 0.283 }               & {\ul 0.556}          & {\ul 0.333}          \\ \hline
\end{tabular}
\caption{Minimum F1 score of classification predictors for DS1.}
\label{tab:classDS1min}
\end{table}

{\ul Observations for DS1:} XGBoost combined with ReliefF displays the best scores for all the settings, as shown in Table \ref{tab:classDS1}. To better assess its superiority, in Table \ref{tab:classDS1min}, we provide the lowest F1 score among the 10 DS1 instances for the best performing configuration of each technique; still the combination XGBoost-ReliefF dominates. Apart from this main observation, several additional interesting observations can be drawn: (i)  For XGBoost, the difference between not employing dimensionality reduction and employing PCA has relatively small difference. More importantly, if ReliefF is not employed, there is a significant degradation in performance when the prediction window is narrower (settings B and D) compared to the settings A and C. (ii) RF cannot handle cases B and D well, while it requires either PCA or ReliefF. (iiii) KNN is superior to RF, especially when there is a buffer window (settings C and D). In addition, KNN worst behavior as shown in Table \ref{tab:classDS1min} is significantly better than RF. (iv) NN techniques exhibit higher F1 scores than RF but lower than XGBoost and KNN in all cases except B. In general, regarding NN flavors, there is not a clear pattern. Employing ReliefF or PCA is always better than not employing any of them. On average, LSTM is never better than MLP or CNN, but its worst performance for setting B is better than the worst performance of the latter two NN architectures. As in all techniques apart from XGBoost-ReliefF, the F1 scores in settings B and D are particularly low. (v) The best average  performing combination for each algorithm (underlined in Table \ref{tab:classDS1})  is not always the one with the better worst behavior (underlined in Table \ref{tab:classDS1min}). Finally, the two dummy classifiers often behave better than many combinations, which is attributed to the high difficulty in training effective models in such settings.  

Further to the last observation, in the following, we do not examine some inferior combinations, such as those not employing PCA or 
ReliefF. We continue the discussion for DS2-DS6, while detailed results are in Tables \ref{tab:classDSall} and \ref{tab:classDSall-min}.


\begin{table}[!]
\small
\resizebox*{!}{\textheight}{%
\begin{tabular}{|c|c|c|c|c|c|}
\hline
\multicolumn{2}{|c|}{Setting}                                                                    & A               & B                  & C                   & D                   \\ \hline
Algorithm                                           & Feat.   Sel.                           & \multicolumn{4}{c|}{Mean F1 score in 10   DS2 instances}                                          \\ \hline 
\multirow{2}{*}{XGBoost}                              & PCA                                             & 0.588                & 0.219                & 0.617                & 0.275          \\ \cline{2-6} 
                                                  & ReliefF                                         & {\ul \textbf{0.701}} & {\ul 0.379}          & {\ul \textbf{0.766}} & {\ul 0.391}    \\ \hline
RF                                               & ReliefF                                             & 0.597                  & 0.291                  & 0.723                 & 0.261 \\ \hline
\multirow{2}{*}{KNN}                              & PCA                                             & {\ul 0.643}          & 0.336                & 0.683                & 0.360          \\ \cline{2-6} 
                                                  & ReliefF                                         & 0.624                & {\ul 0.384}          & {\ul 0.708}          & {\ul 0.396}    \\ \hline
MLP                                               & PCA                                             & 0.627                & 0.428                & 0.669                & \textbf{0.420} \\ \hline
LSTM                                              & ReliefF                                         & 0.566                & 0.298                & 0.649                & 0.369          \\ \hline
\multirow{2}{*}{CNN}                              & PCA                                             & 0.531                & {\ul \textbf{0.467}} & 0.607                & {\ul 0.372}    \\ \cline{2-6} 
                                                  & ReliefF                                         & {\ul 0.536}          & 0.292                & {\ul 0.633}          & 0.366          \\ \hline
\multicolumn{2}{|c|}{Random Classifier}                                                             & 0.484                & 0.225                & 0.519                & 0.282          \\ \hline
\multicolumn{2}{|c|}{All True Classifier}                                                           & 0.459                & 0.174                & 0.494                & 0.190          \\ \hline \hline
Algorithm                                         & Feat.   Selection                             & \multicolumn{4}{c|}{Mean F1 score in 10   DS3 instances}                                          \\ \hline
\multirow{2}{*}{XGBoost}                              & PCA                                             & 0.657                & 0.337                & {\ul 0.694} & 0.368                \\ \cline{2-6} 
                                                  & ReliefF                                         & {\ul \textbf{0.674}} & {\ul \textbf{0.600}} & 0.678                & {\ul \textbf{0.660}} \\ \hline
RF                                               & ReliefF                                             & 0.059                   & 0.036             & 0.038                  & 0.022 \\ \hline
\multirow{2}{*}{KNN}                              & PCA                                             & {\ul 0.660}          & 0.280                & {\ul 0.684}          & 0.339                \\ \cline{2-6} 
                                                  & ReliefF                                         & 0.650                & {\ul 0.597}          & 0.658                & {\ul 0.659}          \\ \hline
MLP                                               & PCA                                             & 0.660                    & 0.515                    & \textbf{0.700}                    & 0.564                    \\ \hline
LSTM                                              & ReliefF                                         & 0.597                    & 0.330                    & 0.658                    & 0.398                    \\ \hline
\multirow{2}{*}{CNN}                              & PCA                                             & 0.564                    & {\ul 0.358}                    &0.612                    & 0.367 \\ \cline{2-6} 
                                                  & ReliefF                                         & {\ul 0.565}                  & 0.351                    &{\ul 0.668 }                   & {\ul 0.388}                  \\ \hline
\multicolumn{2}{|c|}{Random Classifier}                                                             & 0.485                & 0.246                & 0.520                & 0.303                \\ \hline
\multicolumn{2}{|c|}{All True Classifier}                                                           & 0.462                & 0.174                & 0.492                & 0.191                \\ \hline\hline
Algorithm                                         & Feat.  Selection                             & \multicolumn{4}{c|}{Mean F1 score in 10   DS4 instances}                                          \\ \hline
\multirow{2}{*}{XGBoost}                              & PCA                                             & 0.633                & {\ul 0.271}          & {\ul \textbf{0.672}} & 0.297                \\ \cline{2-6} 
                                                  & ReliefF                                         & {\ul 0.653}          & 0.232                & 0.663                & {\ul 0.301}          \\ \hline

RF                                               & ReliefF                                             & 0.058                    & 0.077              & 0.034                   & 0.058 \\ \hline

\multirow{2}{*}{KNN}                              & PCA                                             & 0.611                & 0.229                & 0.665                & 0.274                \\ \cline{2-6} 
                                                  & ReliefF                                         & {\ul \textbf{0.655}} & {\ul 0.305} & {\ul 0.671}          & {\ul 0.339} \\ \hline
MLP                                               & PCA                                             & 0.624                   & \textbf{0.342}                    & 0.652                   & 0.394                    \\ \hline
LSTM                                              & ReliefF                                         &0.616                    &0.314                    &0.643                    & 0.399                    \\ \hline
\multirow{2}{*}{CNN}                              & PCA                                             &0.545                    & 0.304                    &0.590                   &0.347 \\ \cline{2-6} 
                                                  & ReliefF                                         &{\ul  0.565}                   &{\ul  0.313}       &{\ul 0.634 }      &{\ul  \textbf{0.402}}                   \\ \hline
\multicolumn{2}{|c|}{Random Classifier}                                                             & 0.531                & 0.227                & 0.569                & 0.285                \\ \hline
\multicolumn{2}{|c|}{All True Classifier}                                                           & 0.460                & 0.176                & 0.496                & 0.190                \\ \hline \hline
Algorithm                                         & Feat.  Selection                             & \multicolumn{4}{c|}{Mean F1 score in 10   DS5 instances}                                    \\ \hline
\multirow{2}{*}{XGBoost}                              & PCA                                             & 0.560                & 0.245                & 0.628                & 0.275          \\ \cline{2-6} 
                                                  & ReliefF                                         & {\ul \textbf{0.689}} & {\ul 0.351}          & {\ul \textbf{0.752}} & {\ul 0.391}    \\ \hline

RF                                               & ReliefF                                             & 0.618                     & 0.217               & 0.719                    & 0.219 \\ \hline

\multirow{2}{*}{KNN}                              & PCA                                             & 0.627                & 0.322                & 0.688                & 0.369          \\ \cline{2-6} 
                                                  & ReliefF                                         & {\ul 0.640}          & {\ul 0.338}          & {\ul 0.701}          & {\ul 0.388}    \\ \hline
MLP                                               & PCA                                             & 0.611                & 0.418                & 0.666                & \textbf{0.412} \\ \hline
LSTM                                              & ReliefF                                         & 0.575                & 0.293                & 0.663                & 0.364          \\ \hline
\multirow{2}{*}{CNN}                              & PCA                                             & {\ul 0.554}          & {\ul \textbf{0.428}} & {\ul 0.623}          & {\ul 0.367}    \\ \cline{2-6} 
                                                  & ReliefF                                         & 0.504                & 0.388                & 0.572                & 0.342          \\ \hline
\multicolumn{2}{|c|}{Random Classifier}                                                             & 0.484                & 0.225                & 0.519                & 0.282          \\ \hline
\multicolumn{2}{|c|}{All True Classifier}                                                           & 0.459                & 0.174                & 0.494                & 0.190          \\ \hline \hline
Algorithm                                         & Feat.   Selection                             & \multicolumn{4}{c|}{Mean F1 score in 10   DS6 instances}                        \\ \hline
\multirow{2}{*}{XGBoost}                              & PCA                                             & 0.330          & 0.180          & 0.355          & 0.196                \\ \cline{2-6} 
                                                  & ReliefF                                         & {\ul 0.394}    & {\ul 0.333}    & {\ul 0.452}    & {\ul 0.222}          \\ \hline

RF                                               & ReliefF                                             & 0.338                      & 0.240               & 0.351                     & 0.127 \\ \hline

\multirow{2}{*}{KNN}                              & PCA                                             & 0.446          & 0.293          & 0.455          & 0.315                \\ \cline{2-6} 
                                                  & ReliefF                                         & {\ul 0.475}    & {\ul 0.338}    & {\ul 0.489}    & {\ul \textbf{0.319}} \\ \hline
MLP                                               & PCA                                             & \textbf{0.479} & \textbf{0.418} & \textbf{0.492} & 0.310                \\ \hline
LSTM                                              & ReliefF                                         & 0.437          & 0.212          & 0.480          & 0.261                \\ \hline
\multirow{2}{*}{CNN}                              & PCA                                             & 0.408          & {\ul 0.402}    & 0.422          & 0.241                \\ \cline{2-6} 
                                                  & ReliefF                                         & {\ul 0.438}    & 0.257          & {\ul 0.456}    & {\ul 0.294}          \\ \hline
\multicolumn{2}{|c|}{Random Classifier}                                                             & 0.347          & 0.135          & 0.345          & 0.158                \\ \hline
\multicolumn{2}{|c|}{All True Classifier}                                                           & 0.252          & 0.094          & 0.265          & 0.099                \\ \hline
\end{tabular}
}
\caption{Mean F1 score of classification predictors for DS2-DS6.}
\label{tab:classDSall}
\end{table}

\begin{table}[!]
\small
\resizebox*{!}{0.83\textheight}{%
\begin{tabular}{|c|c|c|c|c|c|}
\hline
\multicolumn{2}{|c|}{Setting}                                                                    & A               & B                  & C                   & D                   \\ \hline
Algorithm                                           & Feat.   Sel.                           & \multicolumn{4}{c|}{Min F1 score in 10   DS2 instances}                                          \\ \hline 
\multirow{2}{*}{XGBoost}                              & PCA                                             & 0.462               & 0.095                & 0.514                & 0.154          \\ \cline{2-6} 
                                                  & ReliefF                                         & {\ul 0.564} & {\ul 0.222}          & {\ul 0.634} & {\ul 0.263}    \\ \hline
RF                                               & ReliefF                                             & 0.391                  & 0.08                  & 0.571                 & 0.118 \\ \hline
\multirow{2}{*}{KNN}                              & PCA                                             & {\ul \textbf{0.596}}          & 0.226                & {\ul \textbf{0.652}}                & 0.311          \\ \cline{2-6} 
                                                  & ReliefF                                         & 0.558                & {\ul 0.308}          & 0.629          & {\ul 0.348}    \\ \hline
MLP                                               & PCA                                             & 0.493                & 0.326                & 0.566                & \textbf{0.356} \\ \hline
LSTM                                              & ReliefF                                         & 0.484                & 0.228                & 0.523                & 0.316          \\ \hline
\multirow{2}{*}{CNN}                              & PCA                                             & {\ul 0.487}             & {\ul \textbf{0.345}} &  {\ul 0.563}                & 0.326    \\ \cline{2-6} 
                                                  & ReliefF                                         &  0.484         & 0.219                &0.556          &  {\ul 0.344}          \\ \hline\hline
Algorithm                                         & Feat.   Selection                             & \multicolumn{4}{c|}{Min F1 score in 10   DS3 instances}                                          \\ \hline
\multirow{2}{*}{XGBoost}                              & PCA                                             & {\ul 0.54}                & 0.143                & {\ul 0.606} & 0.25                \\ \cline{2-6} 
                                                  & ReliefF                                         & 0.526 & {\ul 0.200} & 0.512                & {\ul \textbf{0.457}} \\ \hline
RF                                               & ReliefF                                             & 0.0                   & 0.0             & 0.0                  & 0.0 \\ \hline
\multirow{2}{*}{KNN}                              & PCA                                             & 0.576          & 0.148                & {\ul \textbf{0.627}}          & 0.167                \\ \cline{2-6} 
                                                  & ReliefF                                         & {\ul \textbf{0.591 }    }           & {\ul 0.267}          & 0.549                & {\ul 0.45}          \\ \hline
MLP                                               & PCA                                             & 0.529                    & \textbf{0.367}                    & 0.615                    & 0.41                    \\ \hline
LSTM                                              & ReliefF                                         & 0.475                   & 0.233                    & 0.612                    & 0.348                    \\ \hline
\multirow{2}{*}{CNN}                              & PCA                                             & {\ul 0.462}                 &  0.281                    &0.531           & {\ul 0.327}\\ \cline{2-6} 
                                                  & ReliefF                                         & 0.452                 & {\ul 0.282}                    &{\ul 0.579 }                   & 0.314             \\ \hline \hline
Algorithm                                         & Feat.  Selection                             & \multicolumn{4}{c|}{Min F1 score in 10   DS4 instances}                                          \\ \hline
\multirow{2}{*}{XGBoost}                              & PCA                                             & 0.524                & {\ul 0.129}          & {\ul 0.588} & 0.138                \\ \cline{2-6} 
                                                  & ReliefF                                         & {\ul 0.545}          & 0.118                & 0.522                & {\ul 0.2}          \\ \hline

RF                                               & ReliefF                                             & 0.0                 & 0.0              & 0.0                   & 0.0 \\ \hline

\multirow{2}{*}{KNN}                              & PCA                                             & 0.553               & 0.121               & {\ul 0.576  }             & 0.148                \\ \cline{2-6} 
                                                  & ReliefF                                         & {\ul \textbf{0.576}} & {\ul \textbf{0.263}} & 0.554          & {\ul 0.291} \\ \hline
MLP                                               & PCA                                             & 0.517                   & 0.212                    & 0.592                   & 0.299                    \\ \hline
LSTM                                              & ReliefF                                         &0.533                    &0.225                    &0.581                    & 0.338                    \\ \hline
\multirow{2}{*}{CNN}                              & PCA                                             &{\ul 0.474}                    & {\ul \textbf{0.263}}                    &0.521                   &0.286 \\ \cline{2-6} 
                                                  & ReliefF                                         & 0.462                   &0.24       &{\ul \textbf{0.593} }      &{\ul  \textbf{0.37}}                   \\ \hline \hline
Algorithm                                         & Feat.  Selection                             & \multicolumn{4}{c|}{Min F1 score in 10   DS5 instances}                                    \\ \hline
\multirow{2}{*}{XGBoost}                              & PCA                                             & 0.419                & 0.214                & 0.491                & 0.154          \\ \cline{2-6} 
                                                  & ReliefF                                         & {\ul 0.5} & {\ul 0.235}          & {\ul 0.588} & {\ul 0.279}    \\ \hline

RF                                               & ReliefF                                             & 0.408                     & 0.069               & 0.571                    & 0.091 \\ \hline

\multirow{2}{*}{KNN}                              & PCA                                             & {\ul \textbf{0.578}   }             &{\ul  0.222 }               & {\ul \textbf{0.638}}      & 0.27          \\ \cline{2-6} 
                                                  & ReliefF                                         &  0.5          & 0.216          & 0.612          & {\ul 0.339}    \\ \hline
MLP                                               & PCA                                             & 0.514                & 0.333                & 0.571                & \textbf{0.333} \\ \hline
LSTM                                              & ReliefF                                         & 0.508                & 0.22                & 0.557                & 0.296          \\ \hline
\multirow{2}{*}{CNN}                              & PCA                                             & {\ul 0.483}          & {\ul \textbf{0.339}} & {\ul 0.563}          & {\ul 0.304}    \\ \cline{2-6} 
                                                  & ReliefF                                         & 0.286                & 0.267                & 0.351                & 0.0          \\ \hline \hline
Algorithm                                         & Feat.   Selection                             & \multicolumn{4}{c|}{Min F1 score in 10   DS6 instances}                        \\ \hline
\multirow{2}{*}{XGBoost}                              & PCA                                             & {\ul 0.167}      & 0.143          & {\ul 0.167}          & {\ul 0.143}                \\ \cline{2-6} 
                                                  & ReliefF                                         & 0.125    & {\ul 0.19}    & {\ul 0.167}    & 0.0          \\ \hline

RF                                               & ReliefF                                             & 0.1                      & 0.154               & 0.118                     & 0.0 \\ \hline

\multirow{2}{*}{KNN}                              & PCA                                             & 0.25          &{\ul  0.24 }         & {\ul 0.4}          & {\ul 0.194}               \\ \cline{2-6} 
                                                  & ReliefF                                         & {\ul 0.267}    & 0.235    & {\ul 0.308}    & 0.143 \\ \hline
MLP                                               & PCA                                             & \textbf{0.333} & 0.211 & 0.286 & \textbf{0.235 }               \\ \hline
LSTM                                              & ReliefF                                         & 0.25          & 0.093          & 0.345          & 0.175                \\ \hline
\multirow{2}{*}{CNN}                              & PCA                                             & {\ul \textbf{0.333}}          & {\ul \textbf{0.32}}    & 0.246          & 0.195                \\ \cline{2-6} 
                                                  & ReliefF                                         & {\ul \textbf{0.333}}    & 0.146          & {\ul \textbf{0.317}}    & {\ul 0.231}          \\ \hline
\end{tabular}
}
\caption{Minimum F1 score of classification predictors for DS2-DS6.}
\label{tab:classDSall-min}
\end{table}

{\ul Observations for DS2:} The main difference between DS1 and DS2 is that DS2 contains shorter and more diverse patterns before target events.
In average, the combination XGBoost-ReliefF is still the dominant one for settings A and C, which correspond to a larger prediction window.  However, for the settings
B and D where the correct prediction period is  shorter (6 days) the NN solutions, and more specifically CNN and MLP, respectively, perform better than 
XGBoost. Also, NN-based solutions seem to perform better when combined with PCA rather than  ReliefF. Finally, when considering the instance with worst performance, KNN is more robust than XGBoost regarding the A and C settings. 

{\ul Observations for DS3:}  For the DS3, which is similar to DS1 but with a ten-fold increase in the number of event types, the combination of XGBoost-ReliefF is, on average, superior, but (i) the difference in performance compared to KNN and NN-based solutions is smaller, and (ii) in setting C, a NN-based solution, namely MLP, is the best performing one.  While all NN flavors can handle settings A and C in a manner that their performance is not much inferior to the best performing one, only MLP performs well in settings B and D.
As in DS2, KNN exhibits higher robustness with regards to the worst performing instance compared to XGBoost. Finally, RF cannot cope with the challenges of this DS type.

{\ul Observations for DS4:}  DS4 modifies DS2 by examining the behavior after a 10-fold increase in the event types. XGBoost-ReliefF is close to the best performance  for settings A and C but it is not the most efficient anymore. KNN is also very competent for the A and C cases, while NN-based solutions exhibit the highest performance in settings B and D, with MLP-PCA achieving the best scores on average. In terms of robustness, KNN and CNN are the dominant solutions.

{\ul Observations for DS5:}  DS5 introduces shuffled patterns into DS2, which increase the diversity in the patterns preceding target events. The behavior is similar to  DS2 with a slight decrease in the F1 scores overall. This implies that the fact that the warning event patterns contain their elements in random order does not play a key role. Compared to DS4, the behavior is also similar, but the F1 scores are higher, which implies that the techniques are more heavily affected by the increase in the size of the event types.

{\ul Observations for DS6:} DS6 employs the same setting as DS5 but with half training and test instances. This heavily impacts all classifiers but NN-based solutions, and MLP-PCA in particular, exhibit the most limited degradation in their performance. KNN is the second best approach for this DS type.

{\ul Generic observations:}
Overall, the main lesson learned is that, for the classification-based techniques, XGBoost combined with ReliefF is a good starting point, as evidenced also in the proposal in \cite{wang2017predictive}. But there are several other additional dimensions that need to be considered, which extend this main intermediate conclusion: (1) XGBoost alone
does not guarantee good performance; e.g., when combined with PCA, in some settings like D, it cannot even beat a random classifier. (2) RF, which is shown to have good performance in \cite{wang2017predictive} cannot cope with many settings. (3) When the preceding patterns are diverse, in the sense that the constituent elements may have multiple alternatives and the prediction window is relatively narrow, NN-based should be examined. Similarly, KNN-based solutions along with NN-based ones need to be considered in cases where the set of event types is large and/or the failure instances are infrequent.


\begin{table}[tb!]
\begin{tabular}{|c|c|c|c|c|c|}
\hline
\multicolumn{2}{|c|}{Setting}                                                                    & A               & B                  & C                   & D                   \\ \hline
Algorithm                                           & Feat.   Sel.                           & \multicolumn{4}{c|}{Mean F1 score in 10   DS1 instances}                                          \\ \hline
\multirow{3}{*}{XGBoost}                            & None                                          & 0.545                & 0.323                & 0.708                & 0.315                \\ \cline{2-6} 
                                                    & PCA                                           & 0.611                & 0.370                & 0.706                & {\ul 0.409}          \\ \cline{2-6} 
                                                    & ReliefF                                       & {\ul \textbf{0.640}} & {\ul \textbf{0.459}} & {\ul 0.717}          & 0.353                \\ \hline
\multirow{3}{*}{RF}                                 & None                                          & 0.537                & 0.407                & 0.609                & 0.274                \\ \cline{2-6} 
                                                    & PCA                                           & 0.581                & 0.324                & 0.636                & {\ul 0.321}          \\ \cline{2-6} 
                                                    & ReliefF                                       & {\ul 0.630}          & {\ul 0.434}          & {\ul 0.714}          & 0.305                \\ \hline
\multirow{3}{*}{KNN}                                & None                                          & 0.501                & 0.000                & 0.499                & 0.000                \\ \cline{2-6} 
                                                    & PCA                                           & {\ul 0.627}          & {\ul 0.409}          & {\ul 0.700}          & {\ul 0.400}          \\ \cline{2-6} 
                                                    & ReliefF                                       & 0.620                & 0.408                & 0.679                & 0.371                \\ \hline
\multirow{3}{*}{PLS}                                & None                                          & 0.543                & 0.309                & 0.645                & 0.276                \\ \cline{2-6} 
                                                    & PCA                                           & 0.587                & 0.375                & 0.760                & 0.380                \\ \cline{2-6} 
                                                    & ReliefF                                       & {\ul 0.627}          & {\ul 0.433}          & {\ul \textbf{0.764}} & {\ul \textbf{0.415}} \\ \hline
\multirow{3}{*}{MLP}                                & None                                          & 0.506                & 0.150                & 0.576                & 0.259                \\ \cline{2-6} 
                                                    & PCA                                           & 0.505                & {\ul 0.409}          & 0.567                & 0.296                \\ \cline{2-6} 
                                                    & ReliefF                                       & {\ul 0.548}          & 0.402                & {\ul 0.581}          & {\ul 0.306}          \\ \hline
\multirow{3}{*}{LSTM}                               & None                                          & 0.561                & 0.316                & 0.581                & 0.246                \\ \cline{2-6} 
                                                    & PCA                                           & {\ul 0.564}          & 0.305                & {\ul 0.613}          & 0.287                \\ \cline{2-6} 
                                                    & ReliefF                                       & 0.557                & {\ul 0.378}          & 0.605                & {\ul 0.298}          \\ \hline
\multirow{3}{*}{CNN}                                & None                                          & {\ul 0.546}          & 0.294                & 0.543                & 0.245                \\ \cline{2-6} 
                                                    & PCA                                           & 0.502                & 0.309                & 0.546                & 0.291                \\ \cline{2-6} 
                                                    & ReliefF                                       & 0.495                & {\ul 0.333}          & {\ul 0.547}          & {\ul 0.297}          \\ \hline
\multicolumn{2}{|c|}{Random Classifier}                                                             & 0.520                & 0.223                & 0.543                & 0.275                \\ \hline
\multicolumn{2}{|c|}{All True Classifier}                                                           & 0.464                & 0.176                & 0.500                & 0.190                \\ \hline
\end{tabular}
\caption{Mean F1 score of regression predictors for DS1.}
\label{tab:regDS1}
\end{table}

\begin{table}[tb!]
\begin{tabular}{|c|c|c|c|c|c|}
\hline
\multicolumn{2}{|c|}{Setting}                                                                    & A               & B                  & C                   & D                   \\ \hline
Algorithm                                           & Feat.   Sel.                           & \multicolumn{4}{c|}{Min F1 score in 10   DS1 instances}                                          \\ \hline
\multirow{3}{*}{XGBoost}                            & None                                          & 0.442                & 0.207                & 0.515                & 0.285                \\ \cline{2-6} 
                                                    & PCA                                           & 0.483                & 0.323                & 0.613                &  {\ul \textbf{0.381}}         \\ \cline{2-6} 
                                                    & ReliefF                                       & {\ul 0.538}          &{ \ul 0.327} &   {\ul 0.679}          & 0.279 \\ \hline               
\multirow{3}{*}{RF}                                 & None                                          & 0.412                & {\ul 0.333}              & 0.515                & 0.222                \\ \cline{2-6} 
                                                    & PCA                                           & 0.468                & 0.273                & 0.5                & 0.25          \\ \cline{2-6} 
                                                    & ReliefF                                       & {\ul \textbf{0.556}}          & {\ul 0.333}          & {\ul 0.615}          & {\ul 0.276}              \\ \hline
\multirow{3}{*}{KNN}                                & None                                          & 0.378                & 0.000                & 0.355                & 0.000                \\ \cline{2-6} 
                                                    & PCA                                           &{\ul 0.553}                & 0.375                & {\ul 0.679}          & {\ul 0.345} \\ \cline{2-6} 
                                                    & ReliefF                                       &  0.511          & {\ul \textbf{0.378}}          & 0.615                & {\ul 0.345}                \\ \hline
\multirow{3}{*}{PLS}                                & None                                          & 0.427                & 0.273                & 0.5                & 0.222                \\ \cline{2-6} 
                                                    & PCA                                           & 0.448                & 0.327                & 0.68                & 0.35                \\ \cline{2-6} 
                                                    & ReliefF                                       & {\ul 0.5}          & {\ul 0.375}          & {\ul \textbf{0.755}} & {\ul 0.378} \\ \hline
\multirow{3}{*}{MLP}                                & None                                          & {\ul 0.511}           & 0.312                & {\ul 0.555}          & 0.212                \\ \cline{2-6} 
                                                    & PCA                                           & 0.4          & {\ul 0.327 }               & 0.459          & 0.222          \\ \cline{2-6} 
                                                    & ReliefF                                       & 0.416                & 0.222          & 0.486               & {\ul 0.243}                \\ \hline
\multirow{3}{*}{LSTM}                               & None                                          & 0.429                & 0.191                & 0.5                & 0.207                \\ \cline{2-6} 
                                                    & PCA                                           & 0.452                & 0.239                & {\ul 0.526}                & {\ul 0.246}          \\ \cline{2-6} 
                                                    & ReliefF                                       & {\ul 0.455}          & {\ul 0.333}          & 0.507          & 0.22                \\ \hline
\multirow{3}{*}{CNN}                                & None                                          &{\ul  0.455}                & 0.16                & 0.438                & 0.0                \\ \cline{2-6} 
                                                    & PCA                                           & 0.432          & {\ul 0.275}          & {\ul 0.485}                & 0.212              \\ \cline{2-6} 
                                                    & ReliefF                                       & 0.421                &  0.273                & 0.472          & {\ul 0.237}          \\ \hline
\end{tabular}
\caption{Minimum F1 score of regression predictors for DS1.}
\label{tab:regDS1min}
\end{table}

\subsection{Evaluation of Regression Predictors}

We now turn our attention to the regression-based approach that is based on the methodology in \cite{korvesis2018predictive}. The  evaluation of the alternatives follows exactly the same rationale. In all cases, the steepness of the sigmoid function was set to 0.7 and the midpoint to 16 days for settings A and C, and 6 days for settings B and D. The alarm threshold was a configuration parameter, and we present the results corresponding to the best configuration. All the DS instances are the same as previously. The days are grouped in segments of $N=Y$ days, so that again, the counterpart of the observation window is of the same size as the prediction window.

{\ul Observations for DS1:} The average results are presented in Table \ref{tab:regDS1}. The main observations are summarized as follows: (i)
Compared to the results for the classification-based approach in Table \ref{tab:classDS1}, the main conclusion drawn is that the classification-based techniques perform better. Especially for settings B and D, the best performing regression techniques achieve significantly lower F1 scores, while the are marginally superior in setting C. (ii) XGBoost-ReliefF is better in settings A and B, but as shown in Table \ref{tab:regDS1min}, it is not the most robust technique for these settings. In settings C and D, PLS-ReliefF is the dominant solution. (iii) Solutions that employ XGBoost are either the best performing ones or very close to the best performing ones, but without a clear winner regarding dimensionality reduction between PCA and ReliefF. (iv) RF combined with ReliefF can handle settings A, B and C close to the best performing ones and for setting A, it exhibits the highest worst F1 score among the 10 instances tested. (v) KNN is worse than RF in all settings except D and exhibits a consistent pattern, according to which it improves its performance with PCA to a larger degree than with ReliefF. (vi) PLS-RelieF can be deemed as the best solution, since it is superior in C and D and the gap from the superior  technique for settings A and B, XGBoost-ReliefF, is relatively small. (vii) NN-based solutions are clearly inferior to all the afore-mentioned techniques, in the sense that they achieve the lowest average F1 scores in all settings.

\begin{table}[!]
\small
\resizebox*{!}{\textheight}{%
\begin{tabular}{|c|c|c|c|c|c|}
\hline
\multicolumn{2}{|c|}{Setting}                                                                    & A               & B                  & C                   & D                   \\ \hline
Algorithm                                           & Feat.   Sel.                           & \multicolumn{4}{c|}{Mean F1 score in 10   DS2 instances}                                          \\ \hline 
\multirow{2}{*}{XGBoost}                              & PCA                                             & 0.574                & 0.320                & {\ul 0.622}          & {\ul 0.316}          \\ \cline{2-6} 
                                                  & ReliefF                                         & {\ul 0.592}          & {\ul 0.354}          & 0.619                & 0.306                \\ \hline

RF                                               & ReliefF                                         & 0.595                 & 0.329                 & 0.621                 & 0.304                \\ \hline

\multirow{2}{*}{KNN}                              & PCA                                             & 0.617                & 0.326                & {\ul \textbf{0.663}} & {\ul 
\textbf{0.361}} \\ \cline{2-6} 
                                                  & ReliefF                                         & {\ul \textbf{0.634}} & {\ul \textbf{0.363}} & 0.638                & 0.352                \\ \hline
PLS                                               & ReliefF                                         & 0.598                & 0.331                & 0.638                & 0.336                \\ \hline
MLP                                               & ReliefF                                         & 0.522                & 0.337                & 0.568                & 0.326                \\ \hline
LSTM                                              & ReliefF                                         & 0.520                & 0.289                & 0.575                & 0.311                \\ \hline
CNN                                               & ReliefF                                         & 0.478                & 0.277                & 0.534                & 0.296                \\ \hline
\multicolumn{2}{|c|}{Random Classifier}                                                             & 0.484                & 0.225                & 0.519                & 0.282                \\ \hline
\multicolumn{2}{|c|}{All True Classifier}                                                           & 0.459                & 0.174                & 0.494                & 0.190                \\ \hline \hline
Algorithm                                         & Feature   Selection                             & \multicolumn{4}{c|}{Mean F1 score in 10   DS3 instances}                                          \\ \hline
\multirow{2}{*}{XGBoost}                              & PCA                                             & 0.606                & 0.299                & 0.639                & 0.338                \\ \cline{2-6} 
                                                  & ReliefF                                         & {\ul 0.623}          & {\ul \textbf{0.427}} & {\ul 0.657} & {\ul 0.354}          \\ \hline

RF                                               & ReliefF                                         & 0.610                 & 0.383                  & 0.623                  & 0.334            \\ \hline

\multirow{2}{*}{KNN}                              & PCA                                             & {\ul \textbf{0.653}} & 0.349                & 0.654                & 0.400                \\ \cline{2-6} 
                                                  & ReliefF                                         & 0.642                & {\ul 0.373}          & {\ul 0.655}          & {\ul \textbf{0.407}} \\ \hline
PLS                                               & ReliefF                                         & 0.615                & 0.410                & \textbf{0.659}                & 0.366                \\ \hline
MLP                                               & ReliefF                                         & 0.545                   & 0.312                    &0.555                    &0.330                    \\ \hline
LSTM                                              & ReliefF                                         & 0.557                   &0.289                   & 0.572                   &0.311                   \\ \hline
CNN                                               & ReliefF                                         & 0.504                  & 0.271                    &0.523                  &0.304                   \\ \hline
\multicolumn{2}{|c|}{Random Classifier}                                                             & 0.485                & 0.246                & 0.520                & 0.303                \\ \hline
\multicolumn{2}{|c|}{All True Classifier}                                                           & 0.462                & 0.174                & 0.492                & 0.191                \\ \hline \hline
Algorithm                                         & Feature   Selection                             & \multicolumn{4}{c|}{Mean F1 score in 10   DS4 instances}                                          \\ \hline
\multirow{2}{*}{XGBoost}                              & PCA                                             & {\ul 0.626}          & 0.284                & 0.640                & 0.328                \\ \cline{2-6} 
                                                  & ReliefF                                         & 0.625                & {\ul 0.306}          & {\ul 0.644}          & {\ul 0.337}          \\ \hline

RF                                               & ReliefF                                         & 0.612                  & 0.313                   & 0.640                  & 0.322                 \\ \hline

\multirow{2}{*}{KNN}                              & PCA                                             & {\ul \textbf{0.628}} & {\ul \textbf{0.368}} & 0.660                & {\ul \textbf{0.343}} \\ \cline{2-6} 
                                                  & ReliefF                                         & 0.626                & 0.365                & {\ul \textbf{0.664}} & 0.335                \\ \hline
PLS                                               & ReliefF                                         & 0.594                & 0.335                & 0.656                & 0.328                \\ \hline
MLP                                               & PCA                                             & 0.530                    & 0.286                    &0.562                   &0.316                    \\ \hline
LSTM                                              & ReliefF                                         &0.525                   &0.267                  &0.571                   &0.270                    \\ \hline
CNN                                               & ReliefF                                         &0.483                    & 0.260                   &0.533                    & 0.268                  \\ \hline
\multicolumn{2}{|c|}{Random Classifier}                                                             & 0.531                & 0.227                & 0.569                & 0.285                \\ \hline
\multicolumn{2}{|c|}{All True Classifier}                                                           & 0.460                & 0.176                & 0.496                & 0.190                \\ \hline \hline
Algorithm                                         & Feature   Selection                             & \multicolumn{4}{c|}{Mean F1 score in 10   DS5 instances}                                          \\ \hline
\multirow{2}{*}{XGBoost}                              & PCA                                             & 0.581                & 0.314                & {\ul 0.624}          & 0.307                \\ \cline{2-6} 
                                                  & ReliefF                                         & {\ul 0.591}          & {\ul 0.331}          & 0.612                & {\ul 0.322}          \\ \hline

RF                                               & ReliefF                                         & 0.597                   & 0.318                    & 0.607                   & 0.299                 \\ \hline

\multirow{2}{*}{KNN}                              & PCA                                             & 0.604                & 0.326                & {\ul \textbf{0.646}} & {\ul \textbf{0.363}} \\ \cline{2-6} 
                                                  & ReliefF                                         & {\ul \textbf{0.621}} & {\ul \textbf{0.340}} & 0.617                & 0.340                \\ \hline
PLS                                               & ReliefF                                         & 0.605                & 0.321                & 0.637                & 0.328                \\ \hline
MLP                                               & PCA                                             & 0.510                & 0.309                & 0.572                & 0.331                \\ \hline
LSTM                                              & ReliefF                                         & 0.489                & 0.273                & 0.538                & 0.320                \\ \hline
CNN                                               & ReliefF                                         & 0.489                & 0.273                & 0.538                & 0.320                \\ \hline
\multicolumn{2}{|c|}{Random Classifier}                                                             & 0.484                & 0.225                & 0.519                & 0.282                \\ \hline
\multicolumn{2}{|c|}{All True Classifier}                                                           & 0.459                & 0.174                & 0.494                & 0.190                \\ \hline \hline
Algorithm                                         & Feature   Selection                             & \multicolumn{4}{c|}{Mean F1 score in 10   DS6 instances}                              \\ \hline
\multirow{2}{*}{XGBoost}                              & PCA                                             & 0.425          & 0.268                & {\ul 0.446}    & 0.224                \\ \cline{2-6} 
                                                  & ReliefF                                         & {\ul 0.474}    & {\ul \textbf{0.366}} & 0.439          & {\ul 0.242}          \\ \hline

RF                                               & ReliefF                                         & 0.487                    & 0.301                     & 0.452                    & 0.225                 \\ \hline

\multirow{2}{*}{KNN}                              & PCA                                             & {\ul 0.466}    & {\ul 0.330}          & {\ul 0.489}    & {\ul \textbf{0.312}} \\ \cline{2-6} 
                                                  & ReliefF                                         & 0.436          & 0.247                & 0.446          & 0.272                \\ \hline
PLS                                               & ReliefF                                         & \textbf{0.500} & 0.344                & \textbf{0.501} & 0.233                \\ \hline
MLP                                               & ReliefF                                         & 0.394          & 0.300                & 0.415          & 0.250                \\ \hline
LSTM                                              & ReliefF                                         & 0.397          & 0.243                & 0.402          & 0.261                \\ \hline
CNN                                               & ReliefF                                         & 0.355          & 0.242                & 0.373          & 0.268                \\ \hline
\multicolumn{2}{|c|}{Random Classifier}                                                             & 0.347          & 0.135                & 0.345          & 0.158                \\ \hline
\multicolumn{2}{|c|}{All True Classifier}                                                           & 0.252          & 0.094                & 0.265          & 0.099                \\ \hline
\end{tabular}
}
\caption{Mean F1 score of regression predictors for DS2-DS6.}
\label{tab:regDSall}
\end{table}

As previously, for the rest of the DS types, we examine a smaller set of techniques to better focus on the combinations that achieve the highest  F1 scores. The detailed results are shown in Table \ref{tab:regDSall}.

{\ul Observations for DS2:} KNN is the clear winner in this DS type, where the patterns of events preceding a failure become smaller and more diverse. KNN behaves equally well with PCA and ReliefF; the former dimensionality technique is preferable when there is a buffer window (settings C and D). RF and XGBoost behave similarly, while NN-based solutions remain inferior to all other alternatives. In DS2, classification-based solutions perform better than regression-based ones as in DS1.

{\ul Observations for DS3:} The increase of the event types from 150 to 1500 seems to be advantageous for KNN, which along with XGBoost and PLS are the best performing solutions. Still, the regression-based techniques perform worse than their classification counterparts for this DS type, especially for settings B and D.

{\ul Observations for D4:} Similarly to DS2, in DS4 KNN remains the clear winner. In addition, the performance is similar to the one achieved by classification-based solutions.  

{\ul Observations for D5:} As in the previous section, we can observe that shuffling has a small impact and the behavior is very similar to the one for DS2. 

{\ul Observations for D6:} This is a very challenging setting with fewer failure cases for training.  There is no clear winner, but in different settings, any of XGBoost, KNN or PLS may behave better. Also, the performance is slightly better than the one reported in Table \ref{tab:classDSall}.

{\ul Generic observations:} The regression-based methodology is in general inferior to the classification based one in terms of F1 score with an exception in DS6. XGBoost, KNN and PLS are the main options, while NN-based solutions consistently fail to achieve comparable performance with the other solutions. These results extend the remarks originally made in \cite{korvesis2018predictive}, where RF was compared against SVM and was shown to behave better.

\section{Impact of manipulating training data}
\label{sec:train}

\begin{figure}[tb!]
  \centering
  \includegraphics[width=0.5\textwidth]{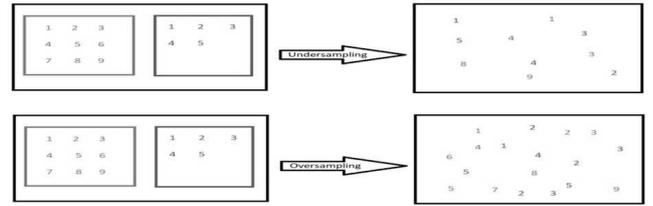}
  \caption{Tackling the class imbalance problem.}
  \label{fig:sampling}
\end{figure}


\begin{table}[tb!]
\begin{tabular}{|c|c|c|c|c|c|c|}
\hline
\multicolumn{3}{|c|}{Setting}                         & A             & B             & C             & D             \\ \hline 
\multicolumn{2}{|c|}{Algorithm} & Feat.   Sel.    & \multicolumn{4}{c|}{Mean F1 score in 10   DS1 instances}                  \\ \hline
\multicolumn{2}{|c|}{XGB}       & ReliefF                & 0.752          & \textbf{0.613} & 0.756          & \textbf{0.609} \\ \hline
\multicolumn{3}{|c|}{Undersampling}                      & \textbf{0.755} & 0.548          & \textbf{0.771} & 0.599          \\ \hline
\multicolumn{3}{|c|}{Oversampling}                       & 0.747          & 0.459          & 0.745          & 0.503          \\ \hline \hline
\multicolumn{3}{|c|}{}                                   & \multicolumn{4}{c|}{Mean F1 score in 10   DS2 instances}                    \\ \hline
\multicolumn{2}{|c|}{XGB}       & ReliefF             & 0.701          & 0.379          & 0.766          & 0.391          \\ \hline
\multicolumn{3}{|c|}{Undersampling}                      & 0.683          & \textbf{0.422} & 0.730          & \textbf{0.416} \\ \hline
\multicolumn{3}{|c|}{Oversampling}                       & \textbf{0.732} & 0.353          & \textbf{0.772} & 0.374          \\ \hline \hline
\multicolumn{3}{|c|}{}                                   & \multicolumn{4}{c|}{Mean F1 score in 10   DS3 instances}                    \\ \hline
\multicolumn{2}{|c|}{XGB}       & ReliefF             & 0.674          & \textbf{0.600} & 0.678          & \textbf{0.660} \\ \hline
\multicolumn{3}{|c|}{Undersampling}                      & 0.656          & 0.565          & 0.671          & 0.592          \\ \hline
\multicolumn{3}{|c|}{Oversampling}                       & \textbf{0.676} & 0.464          & \textbf{0.723} & 0.500          \\ \hline \hline
\multicolumn{3}{|c|}{}                                   & \multicolumn{4}{c|}{Mean F1 score in 10   DS4 instances}                    \\ \hline
\multicolumn{2}{|c|}{MLP}       & PCA                & 0.632          & \textbf{0.337} & 0.662          & 0.382          \\ \hline
\multicolumn{3}{|c|}{Undersampling}                      & \textbf{0.640} & 0.321          & 0.664          & 0.377          \\ \hline
\multicolumn{3}{|c|}{Oversampling}                       & 0.628          & 0.333          & \textbf{0.670} & \textbf{0.386} \\ \hline
\multicolumn{3}{|c|}{}                                   & \multicolumn{4}{c|}{Mean F1 score in 10   DS5 instances}                    \\ \hline \hline
\multicolumn{2}{|c|}{XGB}       & ReliefF            & 0.689          & 0.351          & 0.752          & 0.352          \\ \hline
\multicolumn{3}{|c|}{Undersampling}                      & 0.640          & \textbf{0.398} & 0.739          & \textbf{0.400} \\ \hline
\multicolumn{3}{|c|}{Oversampling}                       & \textbf{0.739} & 0.330          & \textbf{0.768} & 0.351          \\ \hline \hline
\multicolumn{3}{|c|}{}                                   & \multicolumn{4}{c|}{Mean F1 score in 10   DS6 instances}                    \\ \hline
\multicolumn{2}{|c|}{MLP}       & PCA               & \textbf{0.520} & \textbf{0.369} & \textbf{0.491} & \textbf{0.328} \\ \hline
\multicolumn{3}{|c|}{Undersampling}                      & 0.422          & 0.253          & 0.471          & 0.240          \\ \hline
\multicolumn{3}{|c|}{Oversampling}                       & 0.443          & 0.240          & 0.453          & 0.253          \\ \hline
\end{tabular}
 \caption{Results for classification predictors after balancing the training dataset.}
 \label{tab:class-bal}
\end{table}

\begin{table}[tb!]
\begin{tabular}{|c|c|c|c|c|c|c|}
\hline
\multicolumn{3}{|c|}{Setting}                         & A             &B             & C             & D          \\ \hline
\multicolumn{2}{|c|}{Algorithm} & Feat.   Sel.    & \multicolumn{4}{c|}{Mean F1 score in 10   DS1 instances}                  \\ \hline
\multicolumn{2}{|c|}{PLS}       & ReliefF                & \textbf{0.653} & 0.349          & \textbf{0.654} & \textbf{0.400} \\ \hline
\multicolumn{3}{|c|}{Undersampling}                      & 0.629          & \textbf{0.378} & 0.653          & 0.339          \\ \hline
\multicolumn{3}{|c|}{Oversampling}                       & 0.625          & 0.364          & 0.606          & 0.380          \\ \hline \hline
\multicolumn{3}{|c|}{}                                   & \multicolumn{4}{c|}{Mean F1 score in 10   DS2 instances}                    \\ \hline
\multicolumn{2}{|c|}{KNN}       & ReliefF                & 0.634          & 0.363          & 0.638          & 0.352          \\ \hline
\multicolumn{3}{|c|}{Undersampling}                      & \textbf{0.658} & \textbf{0.393} & 0.628          & 0.352          \\ \hline
\multicolumn{3}{|c|}{Oversampling}                       & 0.640          & 0.381          & \textbf{0.650} & \textbf{0.357} \\ \hline \hline
\multicolumn{3}{|c|}{}                                   & \multicolumn{4}{c|}{Mean F1 score in 10   DS3 instances}                    \\ \hline
\multicolumn{2}{|c|}{XGB}       & ReliefF               & \textbf{0.653} & 0.349          & \textbf{0.654} & \textbf{0.400} \\ \hline
\multicolumn{3}{|c|}{Undersampling}                      & 0.629          & \textbf{0.378} & 0.635          & 0.339          \\ \hline
\multicolumn{3}{|c|}{Oversampling}                       & 0.625          & 0.364          & 0.606          & 0.380          \\ \hline \hline
\multicolumn{3}{|c|}{}                                   & \multicolumn{4}{c|}{Mean F1 score in 10   DS4 instances}                    \\ \hline
\multicolumn{2}{|c|}{KNN}       & PCA                & \textbf{0.628} & \textbf{0.368} & \textbf{0.660} & 0.343          \\ \hline
\multicolumn{3}{|c|}{Undersampling}                      & 0.589          & 0.362          & 0.588          & 0.346          \\ \hline
\multicolumn{3}{|c|}{Oversampling}                       & 0.617          & 0.355          & 0.649          & \textbf{0.364} \\ \hline \hline
\multicolumn{3}{|c|}{}                                   & \multicolumn{4}{c|}{Mean F1 score in 10   DS5 instances}                    \\ \hline
\multicolumn{2}{|c|}{KNN}       & ReliefF          & 0.621          & 0.351          & \textbf{0.617}          & 0.340          \\ \hline
\multicolumn{3}{|c|}{Undersampling}                      & \textbf{0.639} & \textbf{0.386} & 0.611          & 0.363          \\ \hline
\multicolumn{3}{|c|}{Oversampling}                       & 0.625          & 0.383          & 0.611          & \textbf{0.369} \\ \hline \hline
\multicolumn{3}{|c|}{}                                   & \multicolumn{4}{c|}{Mean F1 score in 10   DS6 instances}                    \\ \hline
\multicolumn{2}{|c|}{PLS}       & ReliefF             & \textbf{0.500}          & 0.344          & \textbf{0.501} & 0.233          \\ \hline
\multicolumn{3}{|c|}{Undersampling}                      & 0.482          & \textbf{0.367} & 0.435          & \textbf{0.318} \\ \hline
\multicolumn{3}{|c|}{Oversampling}                       & 0.489          & 0.163          & 0.497          & 0.211          \\ \hline
\end{tabular}
 \caption{Results for regression predictors after balancing the training dataset.}
 \label{tab:reg-bal}
\end{table}

The test datasets employed in this work and the vast majority of real-world PdM scenarios in general are inherently characterized by class imbalance and low volume. For example, 
if we have daily data for one year, this corresponds to 365  training examples only. If the observation window exceeds 1 day, as in our settings, the training dataset is decreased even more. Moreover, if the critical failures corresponding to a target event occur 10 times in a year, there are much fewer training samples indicating a failure than those referring to normal operation. Both these factors impact on the methodology efficiency. We have explored two ways to ameliorate the problem of class imbalance, as shown in Figure \ref{fig:sampling}. We either sample the class referring to the normal operation before model building or we oversample the examples referring to failures. After balancing the training dataset, we can oversample the data to increase its size.

The results are shown in Table \ref{tab:class-bal}, where we investigate only the best performing combinations for each dataset type according to the results in the previous section. From the table,  we conclude that in 15 out of the 24 cases (6 datasets X 4 paremeter settings), manipulating the input yielded F1 improvements. Out of these improvements, 7 were through undersampling and 8 through oversampling. The improvements were up to 13.6\% (D setting in DS5). Interestingly, for DS6, in none of the 4 settings manipulating the training dataset has shown to be beneficial.   

We applied the same modifications to the regression-based techniques, and as shown in Table \ref{tab:reg-bal}, there are fewer cases that improvements were observed (14 out of 24). However, the largest relative improvement reached 36.5\% (setting D in DS6). Still, in absolute numbers, this F1 score is inferior to the best performing classification-based technique. For regression-based techniques, inspired by oversampling solutions described in \cite{korvesis2018predictive}, we have explored two additional flavors that perform more systematic undersampling and oversampling based also on the risk values. We omit details, but due to these flavors some additional minor improvements were noticed in DS5 and DS6.

The main lesson learnt is that manipulating the training dataset can indeed yield benefits, but there is no clear pattern regarding its exact usage.  More importantly, as can be easily noticed from the two tables above, in a significant portion of the cases, not only there are no improvements in terms of the F1 score, but the degradation is more significant than the highest observed benefits, e.g., more than 25\%. Therefore, this step needs to be performed with care.

\section{Impact of ensemble solutions}
\label{sec:ensemble}

\begin{figure}[tb!]
  \centering
  \includegraphics[width=0.39\textwidth]{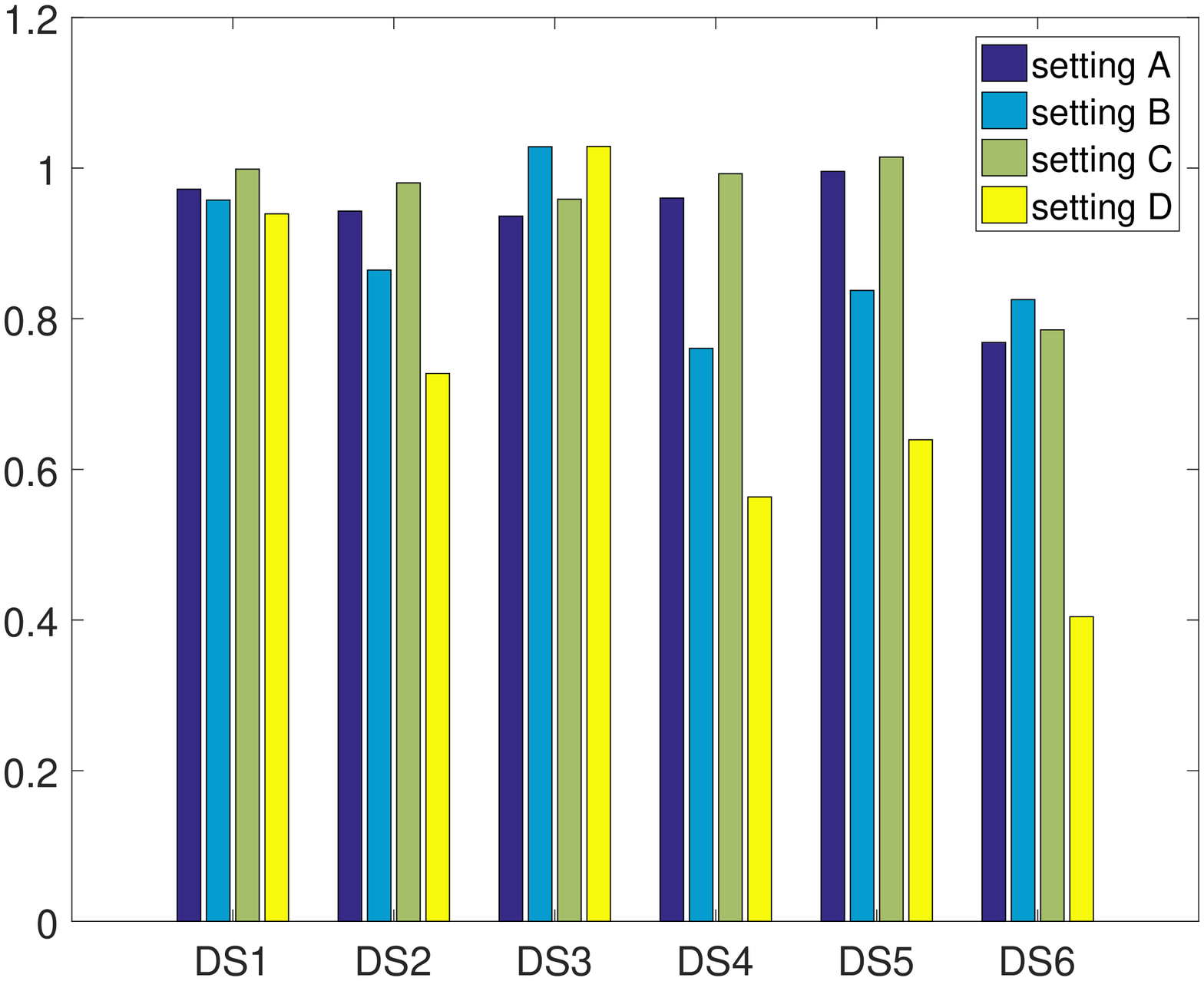} \\
  \includegraphics[width=0.39\textwidth]{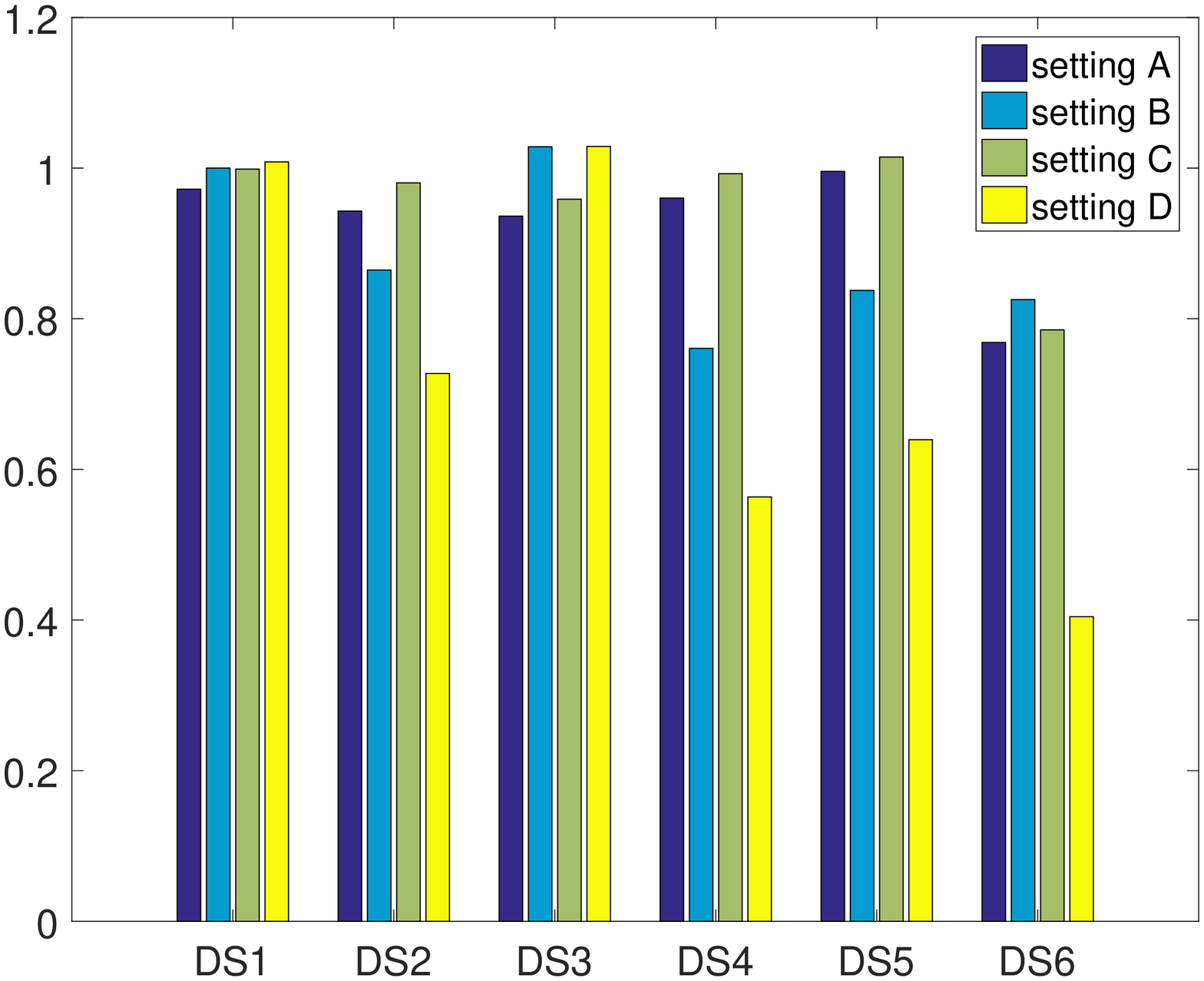} \\
    \includegraphics[width=0.39\textwidth]{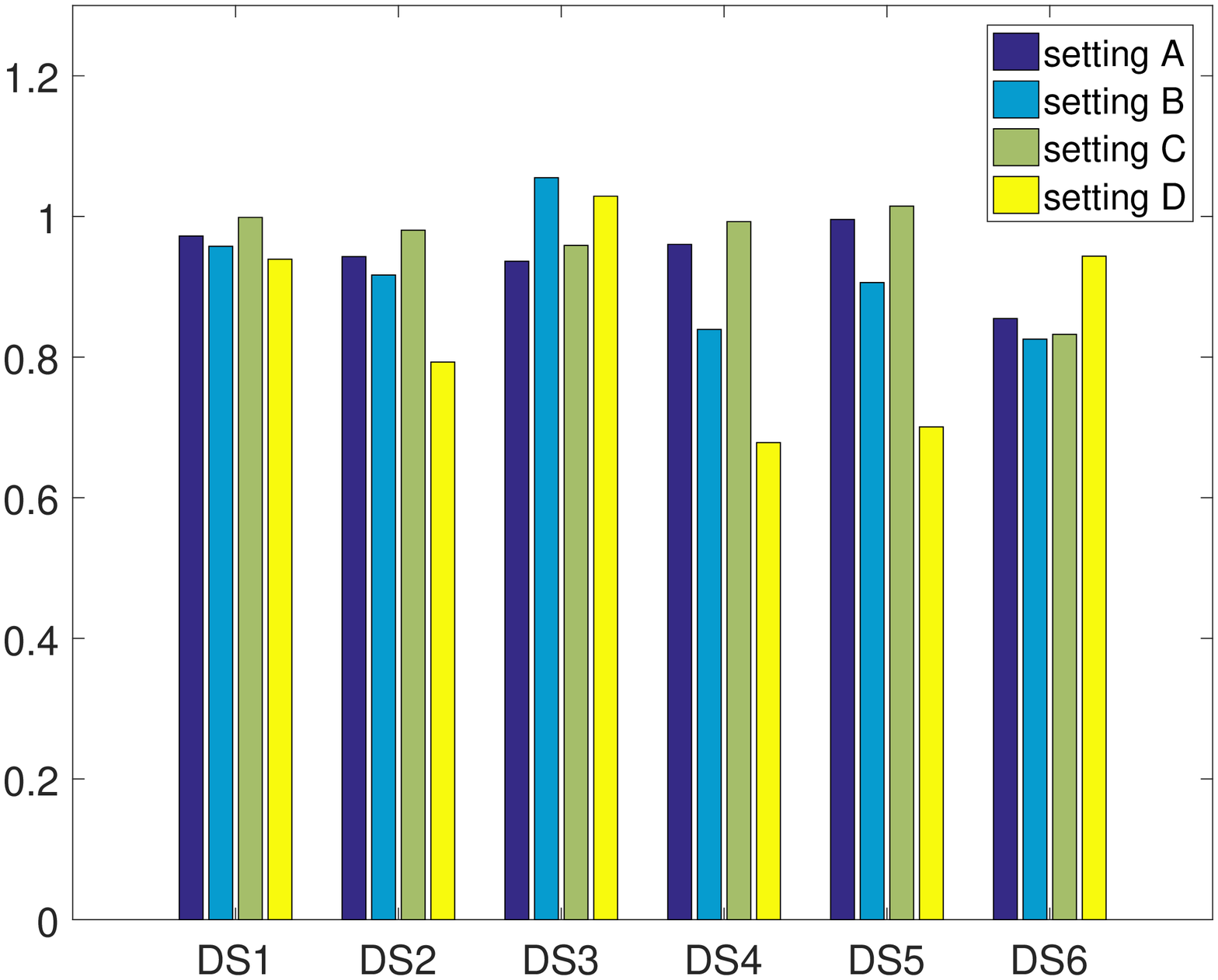} \\
      \includegraphics[width=0.39\textwidth]{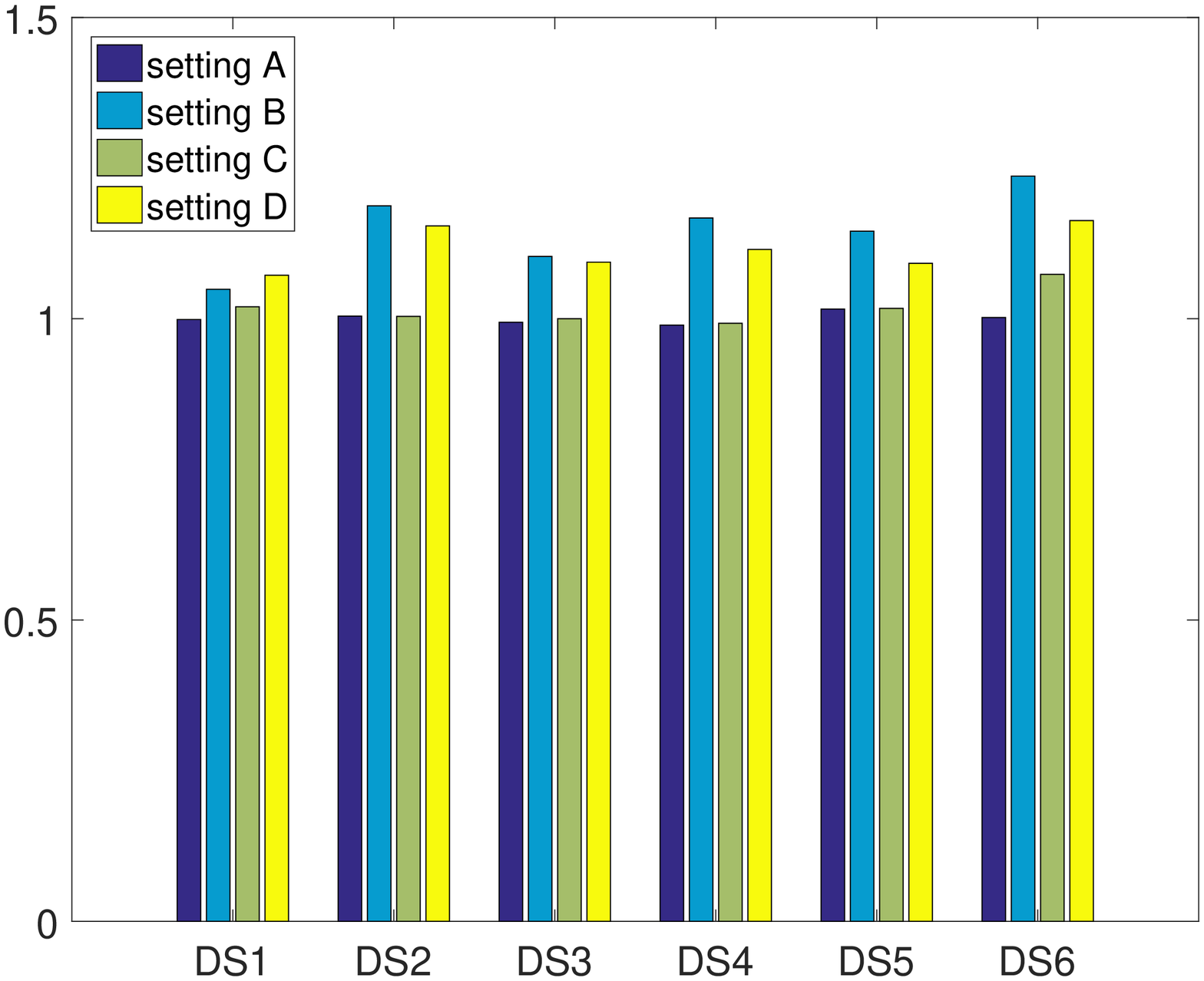} \\  
  \caption{The ratio between the F1 score of the ensemble solution to the best performing single classifier solution for the techniques: a) simple average, b) weighted average, c) dynamic weighted average and d) dynamic threshold (from top to bottom)}
  \label{fig:ensemble}
\end{figure}

The last part of our experimental analysis deals with heterogeneous ensembles. More specifically, we combine multiple classifiers as follows. To narrow the scope of this experiment, we first consider only three classifiers, namely XGBoost, Random Forest and KNN coupled with ReliefF. These techniques are combined in four different manners as follows:
\begin{enumerate}

\item {\bf Simple average}: a prediction for a forthcoming failure is made whenever at least half, i.e, 2 out of 3)  of the classifiers predict so.

\item {\bf Weighted average}: a failure prediction for a forthcoming failure is weighted by the relative F1 score according to the performance of the classifiers in Section \ref{sec:class}. E.g., if the corresponding scores are 0.6, 0.5 and 0.7, the weights become 0.33, 0.28 and 0.39, respectively. If the aggregated value exceeds 0.5, a failure alarm is raised. 

\item {\bf Dynamic weighted average}: the weights here are optimal, in the sense that they are set after each classifier runs according to the classifier performance for each dataset instance; as such, they are different across the 60 datasets. Doing so, we aim to find the upper bound of an ensemble solution based on weights. 

\item {\bf Dynamic threshold}: inspired by our experience in \cite{naskos2019event}, where  in some cases combining solutions with the logical OR instead of AND, we examine a setting where, for each combination of dataset type and parameter setting, we may choose any threshold between 1 and 3 out of 3 classifiers to vote for a failure in order to raise an alarm. In the experiments, we show only the best performing configuration.

\end{enumerate}

Figure \ref{fig:ensemble} shows the results compared to the best performing single-classifier solutions in Section \ref{sec:class}, as reported in Tables \ref{tab:classDS1} and \ref{tab:classDSall}. Simple and weights averaging cannot yield any benefits and lead to performance degradation by 13\% and 12.5\%, respectively. Dynamic weighting can increase the F1 score in some cases, but in general, F1 decreases by 8.2\%. However, dynamic threshold setting manages to produce tangible benefits: the average F1 score improvements are 7.05\%, while, any degradation observed in a combination of dataset type and parameter setting does not exceed 1.05\%.

We repeated a similar experiment for the regression-based predictors. Due to lack of space, we do not present detailed results but the two main observations drawn are: (i) it is beneficial to first combine the  predictors and apply any threshold afterwards, and (ii) the best-performing behavior of the regression-based ensembles  is inferior to the behavior of the worst-performing classification-based ensembles at the top of Figure \ref{fig:ensemble}.

\section{Discussion and Future Work}
\label{sec:concl}

In the introduction, we listed five questions. Here, we provide the summary answers to them, where we group the first three together.

\emph{``What is the impact on the performance of different prediction techniques compared to the ones originally employed or evaluated? Is the behavior of feature selection techniques correlated with the exact prediction technique employed? Which are the most important dataset attributes that affect the technique choices?''}: we have shown that it is beneficial to depart from the initial predictor suggestions in many cases, and also explore KNN- and NN-based solutions. More specifically, our results provide strong evidence that the prediction technique needs to be coupled with the feature selection one, and despite the explosion in the search space, not to treat these aspects separately. The final choice about the technique to be employed depends on several characteristics of the dataset type, including the number of the training examples with failures, the domain size of the event types, and the diversity in the warning patterns. The summary observations are mentioned at the end of Sections 4.3 and 4.4.

\emph{``Does the manipulation of the training set to deal with class imbalance and typically small sizes can yield benefits?''}: we have dealt with this question in Section \ref{sec:train}, where it is shown that improvements can be made. Albeit, there is no clear pattern as to which manipulation techniques should be chosen; therefore a meta-classifier is required, and this constitutes a significant direction for future work.

\emph{``What is the impact of ensemble solutions?''}: as discussed in Section \ref{sec:ensemble}, simply combining predictors in a meaningful way does not guarantee improvements and can even lead to significant performance degradation. The most promising solution is to dynamically set the threshold regarding raising an alarm for a forthcoming failure. This again calls for the development of a meta-classifier, to judiciously take such decisions.

An additional remark from our work is that, in the datasets examined, the classification-based solution outperforms the regression-based one. After reaching this conclusion, further investigation is required to establish the exact cause and the contribution of the extended features extracted from the raw logs in the former technique. 

In summary, in this work we deal with event logs referring to phenomena following Weibull distributions, as is the case in many industrial settings. Assuming that a small portion of such phenomena may be treated as warning signals for a forthcoming significant failure, albeit in a an unclear manner, we have investigated several techniques and methodologies, significantly extending the preliminary results in \cite{NaskosG19} and the conclusions form proposals, such as \cite{wang2017predictive} and \cite{korvesis2018predictive}.

\bibliographystyle{ACM-Reference-Format}
\bibliography{state-of-the-art}

%
% ...

\end{document}